%% file: main.tex
\pdfoutput=1

\documentclass{article}

\usepackage[nonatbib, preprint]{neurips_2023}

\usepackage[utf8]{inputenc} 
\usepackage[T1]{fontenc}    
\usepackage{hyperref}       
\usepackage{url}            
\usepackage{booktabs}       
\usepackage{amsfonts}       
\usepackage{nicefrac}       
\usepackage{microtype}      
\usepackage{xcolor}         

\usepackage{multirow}
\usepackage{makecell}
\usepackage{pifont} 
\usepackage{soul} 
\usepackage{graphicx}
\usepackage{subcaption}
\usepackage{amsmath}
\DeclareMathOperator*{\argmax}{arg\,max}

\title{Zero-Shot Visual Classification with Guided Cropping}

\author{%
  Piyapat Saranrittichai \\
  Bosch Center for Artificial Intelligence \\
  \texttt{Piyapat.Saranrittichai@de.bosch.com} \\
  \And
  Mauricio Munoz \\
  Bosch Center for Artificial Intelligence \\
  \texttt{AndresMauricio.MunozDelgado@bosch.com} \\
  \And
  Volker Fischer \\
  Bosch Center for Artificial Intelligence \\
  \texttt{Volker.Fischer@bosch.com} \\
  \And
  Chaithanya Kumar Mummadi \\
  Bosch Center for Artificial Intelligence \\
  \texttt{chaithanyaKumar.Mummadi@de.bosch.com} \\
}

\begin{document}

\maketitle

\input{sections/00_abstract}

\input{sections/01_introduction}

\input{sections/02_related_works}

\input{sections/03_methodology}

\input{sections/04_experiments}

\input{sections/05_conclusion}


\clearpage

\bibliographystyle{plain}
\bibliography{main.bbl}

\clearpage

\input{sections/07_supplement}

\end{document}

%% file: sections/00_abstract.tex
\begin{abstract}

Pretrained vision-language models, such as CLIP, show promising zero-shot performance across a wide variety of datasets. For closed-set classification tasks, however, there is an inherent limitation: CLIP image encoders are typically designed to extract generic image-level features that summarize superfluous or confounding information for the target tasks. This results in degradation of classification performance, especially when objects of interest cover small areas of input images. In this work, we propose CLIP with Guided Cropping (GC-CLIP), where we use an off-the-shelf zero-shot object detection model in a preprocessing step to increase focus of zero-shot classifier to the object of interest and minimize influence of extraneous image regions. We empirically show that our approach improves zero-shot classification results across architectures and datasets, favorably for small objects.


\end{abstract}


%% file: sections/01_introduction.tex
\section{Introduction}

Conventional supervised learning for closed-set classification tasks involves training Deep Neural Networks (DNNs) on labelled datasets \cite{he2020deep}. The resulting models are inherently limited by the class definitions of a specific task. In contrast, recent research focuses on open-vocabulary zero-shot classification models \cite{jia2021scaling,radford2021learning}. Pretrained with large-scale image-text datasets, these models have more generic class concepts as the definitions can be introduced by textual prompts of natural language.

CLIP is one of the most popular models for open-vocabulary classification \cite{radford2021learning}. Its architecture comprises image and text encoders which encode input images and texts into a shared latent space. These encoders are trained with contrastive losses such that dot product similarity scores between image and text encodings indicate how likely input images and texts correspond to one another.


One limitation of CLIP lies in the fact that its encoders are designed to be generic in the sense that its image encodings encompass entire information of a given image regardless of the target task. While this behavior is desirable for some problems, it simultaneously poses a limitation for closed-set object classification tasks where only certain labels and image contents are of interest. In these cases, encoding entire image contents can lead to suboptimal performance, particularly for small objects. For e.g., in Figure \ref{fig:teaser_full}, the large water region in the image dominates similarity scores between image and text encodings of water-related classes, leading to an incorrect zero-shot prediction.

\input{figures/tex/teaser}

Our central question is: How can we reduce non-discriminative and extraneous information from the image encodings? We observe that reducing areas of context regions by cropping input images around objects of interest can be beneficial. Figure \ref{fig:teaser_crop} illustrates that the cropped image with reduced water regions decrease similarity scores of incorrect water-related classes and result in the dominant similarity score of the correct class (i.e., canoe).

One straightforward approach to reduce influence from non-discriminative information automatically is to directly adopt open-vocabulary object detection models for the zero-shot classification task. These models produce object bounding boxes and \emph{locally} categorize them based on any given text prompts \cite{minderer2022simple,kuo2022f}. However, we speculate that these approaches are not directly optimal for image classification tasks which they are not designed for. In this regard, we conduct an experiment to extend one of the most recent open-vocabulary object detection models, OWL-ViT \cite{minderer2022simple}, for a classification setting where each sample belongs to only one class. We observe that, while OWL-ViT shows reasonable performance on bounding box estimation, its zero-shot classification performance is poor compared to standard zero-shot CLIP baselines (more details in section \ref{sec:exp_ablation_owl_vit}).

In this work, we aim to improve zero-shot object classification performance of CLIP by guiding their focus to the object of interest and reducing the influence of unrelated visual information. Instead of using OWL-ViT for classification directly, we propose to employ it as a bounding box extraction module such that cropped input images are processed by CLIP as shown in Figure \ref{fig:teaser_crop}. We refer this approach as CLIP with Guided Cropping (GC-CLIP). We show that classification performance depends on chosen cropping scales which is especially significant on images with small objects.

Our contributions are as follows: We provide empirical evidence that generic CLIP encoders can lead to suboptimal performance in zero-shot closed-set classification task, particularly on the images with small objects. We propose a method to improve CLIP zero-shot classification using bounding boxes estimated from OWL-ViT. We conduct experiments to show that our approach outperforms a direct OWL-ViT based classifier as well as zero-shot CLIP baselines across different scenarios. Finally, we conduct ablation studies to understand the conditions under which our approach works well.

%% file: figures/tex/teaser.tex
\begin{figure}[t]
    \centering
    \begin{subfigure}[b]{0.45\textwidth}
        \centering
        \includegraphics[width=\textwidth]{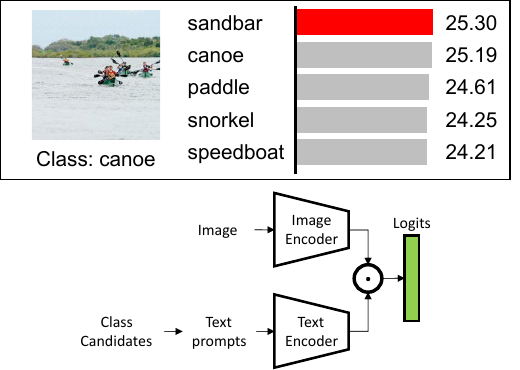}
        \caption{Inference using Conventional CLIP}
        \label{fig:teaser_full}
    \end{subfigure}
    \begin{subfigure}[b]{0.45\textwidth}
        \centering
        \includegraphics[width=\textwidth]{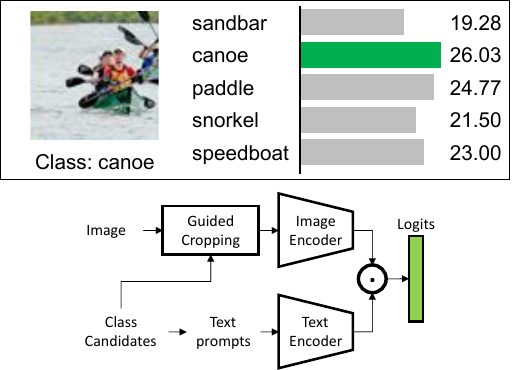}
        \caption{Inference using CLIP with Guided Cropping}
        \label{fig:teaser_crop}
    \end{subfigure}

    \caption{Logits from CLIP (ViT-B/32) before and after cropping around objects of interest}
    \label{fig:teaser}
\end{figure}

%% file: sections/02_related_works.tex
\section{Related Works}

\paragraph{Zero-Shot and Open-Vocabulary Classification}

Zero-shot classification enables trained models to recognize inputs of unseen categories based on externally provided concepts. Earlier works define these concepts in terms of attribute combinations \cite{nagarajan2018attributes,purushwalkam2019task,atzmon2020causal,li2021learning,naeem2021learning,mancini2021open}. However, in open-world applications, it is generally not possible to represent all categories based on limited combinations of trained attributes. Hence, recent research focuses on open-vocabulary classification, in which categories are represented by text prompts. In this regard, images and text prompts can be projected by image/text encoders into a joint embedding space so that their similarities can be computed. CLIP \cite{radford2021learning} and ALIGN \cite{jia2021scaling} encourage similarity between image-text pairs based on contrastive losses. \cite{menon2022visual} improves zero-shot performance by using multiple text prompts per category based on queries from large language models. Florence \cite{yuan2021florence} considers more modalities in addition to images and texts. 

While these models perform well in open-world scenarios, their performance can be limited under the closed-set assumption. As their encoders are designed for open-world applications, they may encode information which are harmful for closed-set classification task. In this work, we aim to alleviate this.



\paragraph{Open-Vocabulary Object Detection} The concept of open-vocabulary has also been investigated in object detection tasks in which object bounding boxes are produced given input text prompts \cite{gu2021open,zhong2022regionclip,li2022grounded,kuo2022f,zhang2022glipv2}. ViLD \cite{gu2021open} trains object detection based on knowledge distillation from pretrained open-vocabulary classification models. In OWL-ViT \cite{minderer2022simple}, simple modifications of standard vision transformers are fine-tuned with large-scale image-text datasets for object detection. GLIPv2 \cite{zhang2022glipv2} extends models to handle various localization tasks.

Object detection models have the innate ability to not only localize, but classify localized objects based on local information. The question may therefore be raised, whether they are in general sufficient to solve the zero-shot classification task alone. In section \ref{sec:exp_ablation_owl_vit}, we conducted experiments based on OWL-ViT, a recent off-the-shelf model, and demonstrate its poor performance on classification tasks. In this work, we use open-vocabulary object detection models only for bounding box extraction.

%% file: sections/03_methodology.tex
\section{Background}

\input{figures/tex/method_overview}

\paragraph{Problem Formulation} Given a test dataset \(\{(x_i, y_i)\}_{i=1}^{N_s} \), where \(x_i \in \mathcal{X} = \mathcal{R}^{w \times w}\) and \(y_i \in \mathcal{Y} = \{1, 2, \ldots, N_c\}\) is an image and its corresponding label, our zero-shot classification task is to construct a prediction function \(F: \mathcal{X} \rightarrow \mathcal{Y} \) based on pretrained open-vocabulary models to maximize the likelihood \(P(\hat{y}|x) = P(F(x)|x)\). Prediction function based on CLIP will be described in this section while our approach will be presented in section \ref{sec:method}.


\paragraph{Conventional CLIP} CLIP \cite{radford2021learning} is a multi-modal model designed for open-vocabulary classification. It consists of an image encoder \(G\) and a text encoder \(H\). To perform closed-set classification, a text prompt \(p^{cls}_j\) needs to be defined for each class \(j \in \mathcal{Y}\). Then, an embedding of each prompt can be obtained by: \(e^{text}_{j} = H(p^{cls}_j)\). During inference, an input image \(x_i\) will be projected into its image embedding \(e^{image}_i = G(x_i)\) so that its classification logit $l^{CLIP}_i$ can be computed as:
\begin{equation}
\label{eq:logit_clip}
l^{CLIP}_i = (E^{text})^T e^{image}_i = \begin{bmatrix} e^{text}_{1} & e^{text}_{2} & \ldots & e^{text}_{N_c} \end{bmatrix}^{T}e^{image}_i.
\end{equation} 
Each entry \(l^{CLIP}_{ij}\) of the logit indicates the similarity score between the (embedded) input image and the \(j\)-th prompt. The final class prediction can then be obtained as \(\hat{y}_i = \argmax_{j \in \mathcal{Y}} l^{CLIP}_{ij}. \)

Above, we assume that one prompt is available per class. However, it has been shown recently that using multiple prompts per class can improve performance \cite{menon2022visual}. In this case, each \(e^{text}_{j}\) from equation \ref{eq:logit_clip} can be replaced with the average embedding computed from all available text prompts of class \(j\).

\section{Methodology}
\label{sec:method}

\subsection{CLIP with Guided Cropping}
\label{sec:method_owl_clip}

Conventionally, image embedding \(e^{image}_{i}\) is computed directly from the full image \(x_i\) without any task-specific constraints. For closed-set classification, especially in cases of a small object image, this implies that potentially unrelated information is also encoded into \(e^{image}_{i}\), which may lead to suboptimal performance. Minimizing the amount of unrelated concept information in image embeddings is desirable in this case. Our approach, CLIP with Guided Cropping (GC-CLIP), achieves this by using bounding box estimates provided by OWL-ViT.

OWL-ViT is an open-vocabulary object detection model \cite{minderer2022simple}. It takes an image and text prompts of target classes as inputs and produces outputs as a set of bounding boxes together with their scores and classes. In this work, we only use OWL-ViT as a bounding box extraction module as its class predictions are not accurate enough (see section \ref{sec:exp_ablation_owl_vit}). The overall GC-CLIP pipeline is shown in Figure \ref{fig:method_overview_computing_boxes}. We only consider top-k classes (we use k=5) to refine the preliminary CLIP predictions. This is reasonable since it has high probabilities that these top-k classes contain the correct class (see appendix \ref{sec:appendix_top_k}).

\paragraph{Candidate box extraction} We detect bounding boxes of each top-k class with OWL-ViT independently. We found that this is more robust to misdetection resulting in better performance compared to detecting bounding boxes of all classes at once (see appendix \ref{sec:appendix_object_detection_pass}). Formally, a set of bounding box candidates \(B_i\) for an image \(x_i\) can be obtained based on OWL-ViT as follows:
\begin{equation}
\label{eq:bounding_box_candidates}
    B_i = \bigcup_{j \in J^k_i} b_{ij}= \bigcup_{j \in J^k_i}OWL(x_i, p^{det}_j)
\end{equation}
where \(J_k \subseteq \mathcal{Y} \) is a set of top-k classes with respect to \(l^{CLIP}_i\), \(p^{det}_j\) is a text prompt for detection of class \(j\) and \(OWL\) is OWL-ViT detection function returning a max-score bounding box with respect to an input image and a prompt. All bounding boxes are adjusted to squares to avoid skewing images when they are, afterward, transformed into a CLIP-compatible image size. (e.g., \(224 \times 224\)). 

\input{figures/tex/gc_clip_box}

\paragraph{Box selection} Next, we need to pick one bounding box from \(B_i\). We start from a primary box \(b^{0}_i \in B_i\) which has the highest estimated score from OWL-ViT. In our experiments, we found that using the primary box directly is generally suboptimal as its crop may be too tight to target objects. It is therefore beneficial to slightly enlarge the box (see section \ref{sec:exp_ablation_margin_vs_acc}). Given \(b^{0}_i\) has the width of \(w_{b^{0}_i}\) and \(x_i\) has the width of \(w\), the box is enlarged to an $\alpha$-margin box \(b^{\alpha}_i\) uniformly in all direction to the size of \(w_{b^{0}_i} + \alpha(w-w_{b^{0}_i})\), where \(\alpha \in [0, 1]\) is called margin ratio (see Figure \ref{fig:gc_clip_box_fix_scale}). For the enlargement, if a box edge exceeds image boundary in one direction, the enlargement will be compensated in the opposite direction. In cases with box augmentation, multiple \(\alpha\) can be employed (see section \ref{sec:method_box_augmentation}).


\paragraph{Logit computation} This selected box \(b^{\alpha}_i\) is used to crop \(x_i\) and resize it to a CLIP-compatible image size \(w \times w\) resulting in a preprocessed image \(x^{\alpha}_i\). The new top-k logit \(l^{GC\_CLIP(k)}_i\) is computed based on \(x^{\alpha}_i\) as follows:
\begin{equation}
\label{eq:logit_gc_clip}
l^{GC\_CLIP(k)}_i = \begin{bmatrix} e^{text}_{j^1} & e^{text}_{j^2} & \ldots & e^{text}_{j^k} \end{bmatrix}^{T} G(x^{\alpha}_i),
\end{equation}
where \(j^1, j^2, \ldots, j^k \in J^k_i\). The final class prediction is the class within \(J^k_i\) corresponding to the maximum entry of \(l^{GC\_CLIP(k)}_i\).

\subsection{Test-Time Box Augmentation}
\label{sec:method_box_augmentation}

\input{figures/tex/clip_noise}

While prediction can directly perform on a raw/preprocessed input image, this can lead to noisy prediction from CLIP. Small non-semantic changes in images can cause changes in predictions making CLIP outputs difficult to analyze. We show this behavior by processing 10 random crops (90\%-100\% of the original widths) of the same image with CLIP. One would expect that, standard deviations of its predicted true-label probabilities should be low and its final class predictions should not change across different crops. However, we notice from Figure \ref{fig:clip_noise_std} that the standard deviations can be relatively high (around 0.2), while the average true-label probability is 0.55. In addition, only around 60\% of test samples have no changes in final class predictions across crops (see Figure \ref{fig:clip_noise_change_cnt}). These results indicate significant sensitivity of CLIP to non-semantic changes. Therefore, instead of computing logits from raw/preprocessed images only, we can perform a simple test-time augmentation to help mitigate this issue. In this work, we investigate two augmentation strategies.

\paragraph{Random Crop Box Augmentation (RAug)} With RAug, we augment a single input (raw or preprocessed) image into \(N_{aug}\) total images by cropping the input image with \(N_{aug}\) boxes of random widths within \([\beta w, w]\), while \(\beta \in (0, 1) \). The augmented images are used to compute multiple predicted logits as per equation \ref{eq:logit_gc_clip}, which can then be averaged to produce the final logit score.

\paragraph{Multi-Margin Box Augmentation (MAug)} In some cases, it is beneficial to consider context information as long as it does not dominate object information. With MAug, we need to firstly obtain the primary box \(b^0_i\). Then, instead of using a margin ratio \(\alpha\) as in section \ref{sec:method_owl_clip}, we perform an object-centric augmentation by using \(N_{aug}\) bounding boxes obtained from multiple margin ratios, distributed uniformly from 0 to 1 (see Figure \ref{fig:gc_clip_box_multi_scale}). In other words, the set of all final boxes used in this augmentation is \(\left\{ b^{\alpha_k}_i | \alpha_k = \frac{k}{N_{aug}-1}, k \in \left\{0, 1, \ldots, N_{aug}-1\right\}  \right\}  \). Similarly, logits computed from images cropped by these final boxes are then averaged to get the final logit score.

It must be noted that, with MAug, regions close to the target object are covered by more boxes compared to regions far from the object. Therefore, the augmentation allows some context information to be considered but with lower importance compared to object information.



%% file: figures/tex/method_overview.tex
\begin{figure*}[tb]

    \centering
    \includegraphics[width=\textwidth]{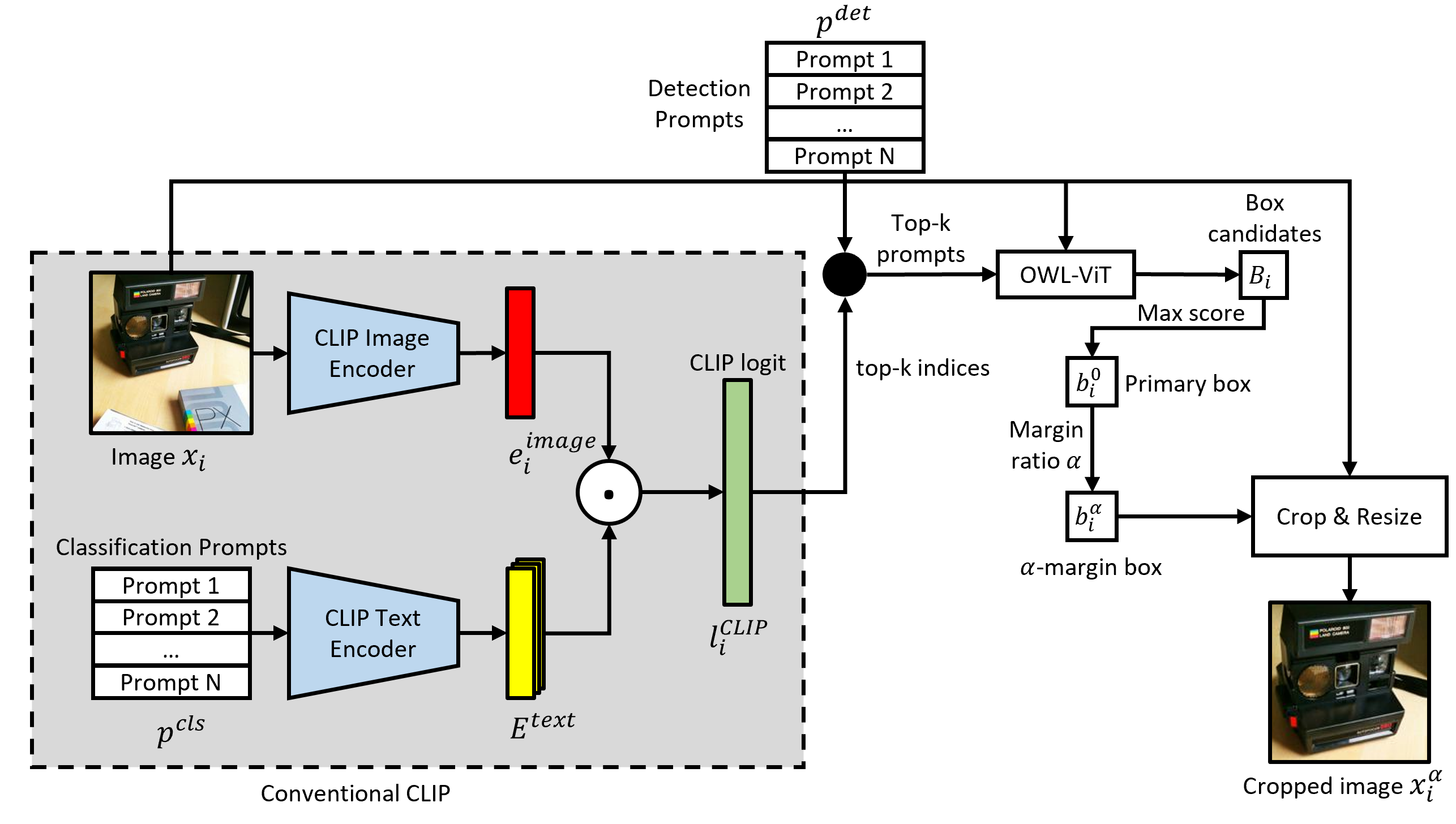}
    \caption{Guided Cropping pipeline to obtain a guided cropped image with margin ratio \(\alpha
    \)}
    \label{fig:method_overview_computing_boxes}

\end{figure*}

%% file: figures/tex/gc_clip_box.tex
\begin{figure*}[tb]
    \centering
    \begin{subfigure}[b]{0.305\textwidth}
        \centering
        \includegraphics[width=\textwidth]{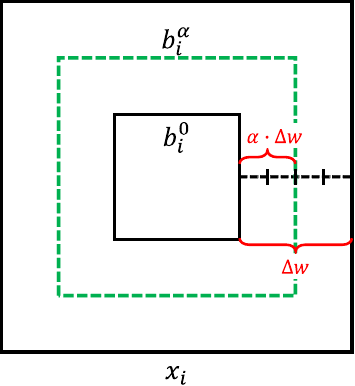}
        \caption{Without augmentation}
        \label{fig:gc_clip_box_fix_scale}
    \end{subfigure} \hspace{2cm}
    \begin{subfigure}[b]{0.38\textwidth}
        \centering
        \includegraphics[width=\textwidth]{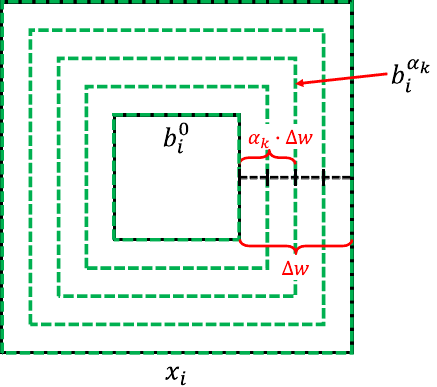}
        \caption{With Multi-Margin augmentation}
        \label{fig:gc_clip_box_multi_scale}
    \end{subfigure}

    \caption{Each green square corresponds to a final bounding box \(b^{\alpha}\) (or \(b^{\alpha_k}\)) which will be used to crop the original image \(x_i\) to produce logit for the final prediction. \(\Delta w\) is the width difference between the original image and the primary box \(b^0_i\). \(\alpha\) and \(\alpha_k\) are margin ratios.}
    \label{fig:gc_clip_box}
\end{figure*}


%% file: figures/tex/clip_noise.tex
\begin{figure*}[tb]
    \centering
    \begin{subfigure}[b]{0.45\textwidth}
        \centering
        \includegraphics[width=\textwidth]{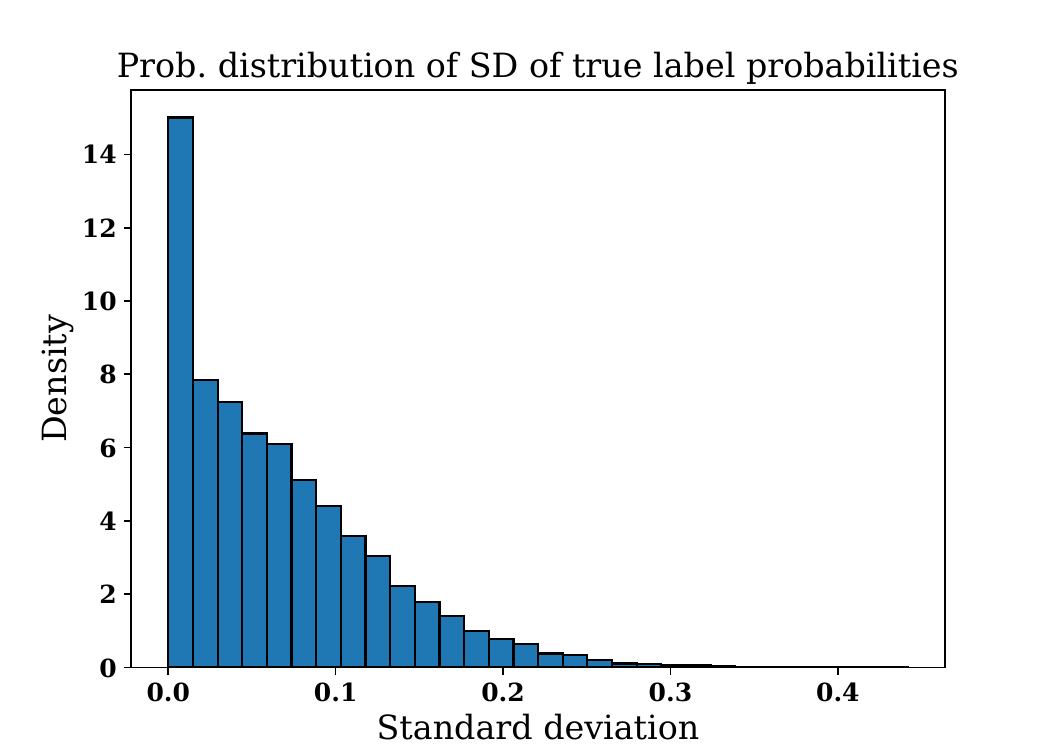}
        \caption{SD of predicted true-label probabilities}
        \label{fig:clip_noise_std}
    \end{subfigure}
    \begin{subfigure}[b]{0.45\textwidth}
        \centering
        \includegraphics[width=\textwidth]{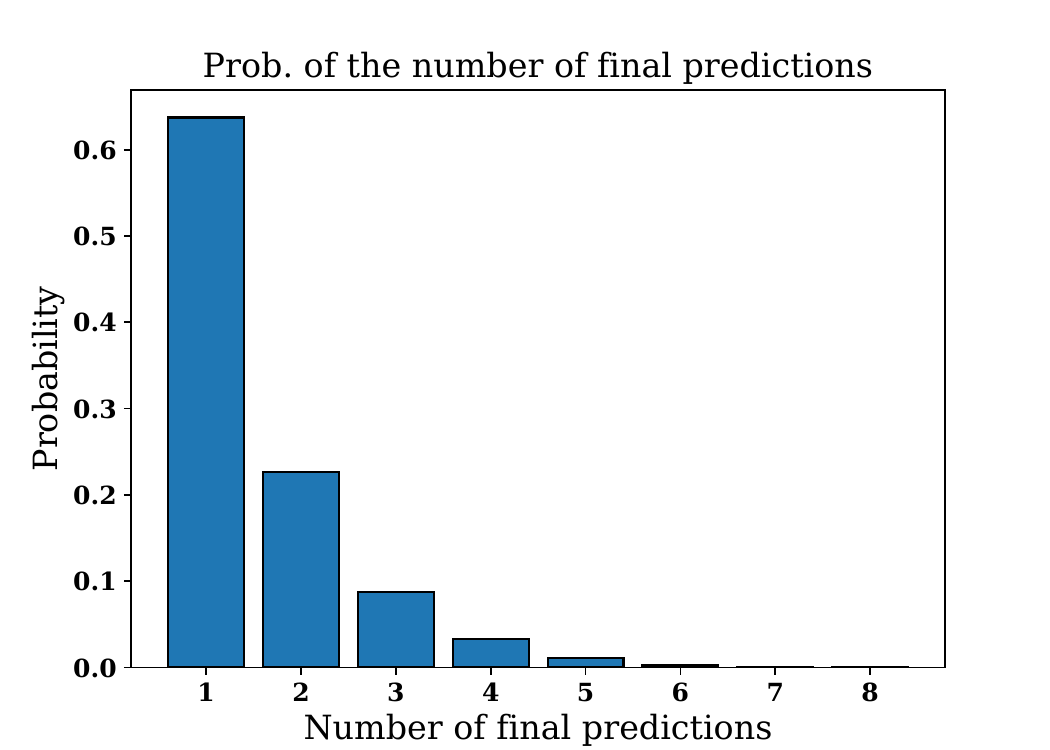}
        \caption{Number of final class predictions across crops}
        \label{fig:clip_noise_change_cnt}
    \end{subfigure}

    \caption{Results when forwarding multiple random crops of the same images (from ImageNetS919 dataset) to CLIP (ViT-B/32) demonstrating CLIP sensitivity to non-semantic changes.}
    \label{fig:clip_noise}
\end{figure*}


%% file: sections/04_experiments.tex
\section{Experiments}
\label{sec:exp}

In this section, we conduct experiments to demonstrate that utilizing CLIP with Guided Cropping can improve zero-shot classification performance. In addition, several ablation studies are also conducted to understand its failure modes and the conditions under which our approach works well.

\subsection{Setup}
\label{sec:exp_setup}

\paragraph{Datasets} We would like to study classification scenarios in which object sizes in images are controllable. In this work, two datasets are employed. (1) ImageNetS \cite{gao2022large}: this dataset is an extension of ImageNet \cite{russakovsky2015imagenet} and originally designed for unsupervised semantic segmentation. We use the validation split of the dataset in which pixel-wise segmentation annotations are available. It contains 12,419 samples of 919 classes in total. We construct a subset with target objects of small sizes, called ImageNetS919-SM, containing 2,334 samples whose object sizes are no more than 20\% of the full image size. (2) CUB \cite{welinder2010caltech}: this dataset is a benchmark for fine-grained classification consisting of 200 bird types. We evaluate our models on its test split of 5,794 samples. Similarly, based on bounding box annotations of the dataset, we construct its subset whose target object sizes are less than 20\% of the full image size resulting in CUB-SM containing 1,390 samples. More details of our dataset splitting and example images of these datasets can be found in the appendix \ref{sec:appendix_small_dataset_construction}.

\paragraph{Baselines} CLIP \cite{radford2021learning} is used as the main architecture of all baselines. We conduct experiments with two classification prompt types similar to \cite{menon2022visual} (1) Category: Each class has a single prompt of its category name (2) Descriptions: Each class has multiple prompts queried automatically from GPT-3 according to \cite{menon2022visual}. In the latter case, the final logit value for a given class is computed by averaging the logit values obtained from all prompts for that class.

\paragraph{Implementation} We apply our Guided Cropping and box augmentation on top of each baseline. For Guided Cropping variations, the margin ratio \(\alpha\) of 0.2 is used unless otherwise specified. We perform box augmentation with \(N_{aug}=11\). For RAug, \(\beta=0.9\) is used. The high value of \(\beta\) makes RAug augmented boxes less likely to crop object contents away. CLIP backbones studied in this work are ViT-B/32, ViT-B/16 and ViT-L/14. For OWL-ViT, its backbone is ViT-B/32 for all experiments. Category names are used as prompts to perform detection with OWL-ViT. The code of our implementation will be publicly available upon paper acceptance.

\subsection{Zero-Shot Classification Performance}
\label{sec:exp_overall_performance}

\input{tables/far_performance}

In this section, we evaluate zero-shot classification performance of different model configurations on various datasets including both unconstrained object sizes (full dataset) and small-object variants (with -SM suffix). The results are shown in Table \ref{table:far_performance}.

Considering datasets with unconstrained object sizes, ImageNetS919 and CUB, our Guided Cropping performance is generally comparable to (or slightly better than) non-Guided Cropping baselines. This is expected since many samples in these cases could have objects whose sizes already dominate the scene. On the other hand, both box augmentations consistently improve classification performance in all cases indicating that raw predictions from CLIP models are indeed noisy. Smoothing their predictions with box augmentations helps our methods to be more robust to this noise.

Considering results on datasets with small object sizes, ImageNetS919-SM and CUB-SM, our Guided Cropping demonstrates consistent improvement over baselines across different model configurations. This trend can also be noticed regardless of the prompt types. This indicates that our approach, as expected, is more beneficial for images with small target objects. This is reasonable since small object images leave more space in the images for context information which should be reduced before performing image encoding. Another interesting observation is that employing GC-CLIP with Multi-Margin augmentation (MAug) generally achieved better performance. This infers that hinting the context cues with lower importance can complement with the focus on object of interest to make definite and correct decisions.

It must be noted that, in this experiment, we integrate our Guided Cropping on top of zero-shot models. A question may arise: how does our Guided Cropping affect pretrained supervised models? We conduct an experiment and found that pretrained supervised models benefit less from cropping with small bounding boxes (see appendix \ref{sec:appendix_supervised}). This is expected since supervised models can exploit unrelated contexts as shortcuts \cite{geirhos2020shortcut} to gain performance on in-distribution samples.

\subsection{Importance of Margin Ratio}
\label{sec:exp_ablation_margin_vs_acc}

\input{figures/tex/margin_vs_acc_far}

Margin ratio (\(\alpha\)) mentioned in section \ref{sec:method_owl_clip} controls how much primary boxes from OWL-ViT are enlarged before they are used to crop input images. Varying margin ratios can help us understand how CLIP reacts to Guided Cropping from \(\alpha=0.0\) (crop with a raw OWL-ViT box) to \(\alpha=1.0\) (no Guided Cropping at all). In this section, we study our models with different margin ratios on ImageNetS919-SM. The results are shown in Figure \ref{fig:margin_vs_acc_far}. We mainly discuss results from GC-CLIP and GC-CLIP+RAug here as these configurations utilize a single margin ratio.

According to the results, when Guided Cropping is applied (\(\alpha < 1\)), classification accuracies are generally better than the accuracies without Guided Cropping (\(\alpha = 1\)). This confirms the benefit of GC-CLIP. It must be noted that, there are some consistent drops of the performance when the values of \(\alpha\) are too small (e.g., when \(\alpha \in [0.0, 0.1]\)). This infers that too tight bounding boxes can degrade classification performance. One explanation of this observation is that, in order to recognize an object, models need to know the object shape clearly. Too tight bounding boxes can make the models having unclear information on the object boundaries leading to performance drops.


\subsection{Understanding Object Size Conditions}
\label{sec:exp_osize_conditions}

\input{figures/tex/object_size_vs_acc_vitb32}

In section \ref{sec:exp_overall_performance}, we only conduct experiments on small object images with only one object size conditions (i.e., maximum relative object sizes \(< 20\%\) of the total image areas). In this section, we would like to explore how our approach performs on different object size conditions. Therefore, we vary maximum relative object sizes of ImageNetS919 dataset from 5\% to 100\% for our evaluation. Details of the samples in individual conditions are given in appendix \ref{sec:appendix_small_dataset_construction}.

The results are shown in Figure \ref{fig:object_size_vs_acc_vitb32} (see appendix \ref{sec:appendix_osize_vs_acc} for the results of other backbones). Considering the cases without any object size constraints (i.e., x-axis = 1.0), applying Guided Cropping does not significantly impact the performance (the same observation in Table \ref{table:far_performance}). However, as the maximum object sizes decrease, accuracy gaps between conventional CLIP and GC-CLIP become larger. The gaps are also more significant when MAug is applied for box augmentation instead of RAug. This experiment highlights conditions with small objects that our approach works well.

\subsection{Qualitative Evaluation}

\input{figures/tex/qualitative_visualization}

In this section, we quantitatively evaluate GC-CLIP by visualizing some samples whose predictions are changed from CLIP. Improved samples are shown in Figure \ref{fig:qualitative_visualization_good}. Reasonable improvements can be noticed among these samples. For example, in the ship image, land and sea are context covering large regions. Considering these contexts excessively makes standard CLIP incorrectly predicting the target object as an amphibious vehicle. However, GC-CLIP recognizes the image focusing on the primary box at the vehicle. This reduces distracted visual information when encoding the image leading to correct prediction.

On the other hand, image samples whose predictions are incorrectly changed by GC-CLIP are shown in Figure \ref{fig:qualitative_visualization_bad}. These samples are failed potentially due to distance between target objects and important contexts. While MAug augmentation allows some contexts to be considered during prediction, large distance between target objects reduce importance of the contexts for the model (less boxes cover the contexts). For example, considering the space shuttle image, the target object is too tiny so ground is an important context distinguishing a missile and a space shuttle (which is usually launched vertically). However, large distance between the ground and the object box reduces effects from the ground in GC-CLIP. Strategies to weight contexts dynamically can be investigated in future works.

\subsection{Can we use OWL-ViT directly as a classifier?}
\label{sec:exp_ablation_owl_vit}

\input{figures/tex/owl_incorrect_samples}

Theoretically, OWL-ViT also has capability to minimize information outside target object boundaries and can be used in zero-shot classification task. In this section, we would like to show that, when OWL-ViT is adopted as a classifier directly, it still has limited performance on our classification task.

In order to use OWL-ViT as a classifier, we need to transform its outputs from sets of bounding box locations, scores and class labels into class-wise logits. In this regard, given an input image, prediction logit of a class can be obtained as follows: Firstly, we iterate whether there are any bounding boxes exist for that class. If any boxes exist, the class logit value will be assigned as the maximum score of its corresponding bounding boxes. Otherwise, its logit will be zero. This simple extension encourages classes of bounding boxes with high scores to have high logits.

We evaluate this classifier on ImageNetS919 dataset and obtain 20.34\% and 40.78\% as top-1 and top-10 accuracies respectively. Here, the performance is still much lower compared to our baseline performance in Table \ref{table:far_performance} indicating poor classification accuracy of this classifier. 

The poor performance of this classifier can be investigated by visualizing incorrectly predicted samples in Figure \ref{fig:owl_incorrect_samples}. While OWL-ViT gives reasonable bounding boxes, its class predictions are inaccurate. The actual classes are likely to be confused with other classes with fine-grained differences. For example, the model misclassifies an image of a tiger shark as a snoek fish whose shape is indeed closely resemble to shark. This significant degradation from fine-grained details confirms that OWL-ViT is not optimal to be used as a classifier on standard classification benchmarks.

%% file: tables/far_performance.tex
\begin{table}[t]

\caption{Zero-shot classification accuracies from different datasets and model configurations.}

\centering
\fontsize{8}{11}\selectfont
\begin{tabular}{c|c|c|c|cc|cc}
\Xhline{4\arrayrulewidth}
\multirow{2}{*}{Model} & \multirow{2}{*}{Prompt} & \multirow{2}{*}{\makecell{Guided\\Cropping}} & \multirow{2}{*}{\makecell{Box Aug.}} & \multicolumn{4}{c}{Dataset} \\ 
 &&&& ImageNetS919 & CUB & ImageNetS919-SM & CUB-SM \\ \hline
\parbox[t]{2mm}{\multirow{10}{*}{\rotatebox[origin=c]{90}{ViT-B/32}}}
  & \multirow{5}{*}{Category}     & -         & -    & $63.62$ & $51.83$ & $52.83$ & $49.57$ \\
 &      & -         & Random Crop & $64.42$ & $52.45$ & $53.47$ & $50.79$ \\
 &      & \ding{51} & -    & $63.61$ & $52.40$ & $55.18$ & $51.44$ \\
 &      & \ding{51} & Random Crop & $64.46$ & \textbf{53.12} & \textbf{56.00} & $52.81$ \\
 &      & \ding{51} & Multi-Margin & \textbf{64.66} & \textbf{53.12} & \textbf{56.00} & \textbf{53.09} \\
 \cline{2-8}
 & \multirow{5}{*}{Descriptions} & -         & -    & $68.54$ & $53.05$ & $55.70$ & $50.14$ \\
 &  & -         & Random Crop & $69.15$ & $53.62$ & $57.33$ & $50.79$ \\
 &  & \ding{51} & -    & $68.59$ & $54.07$ & $58.61$ & \textbf{53.38} \\
 &  & \ding{51} & Random Crop & $69.07$ & $54.47$ & $59.08$ & 53.09 \\
 &  & \ding{51} & Multi-Margin & \textbf{69.62} & \textbf{54.56} & \textbf{60.07} & $52.95$ \\
 \hline
 \hline
 \parbox[t]{2mm}{\multirow{10}{*}{\rotatebox[origin=c]{90}{ViT-B/16}}}
 & \multirow{5}{*}{Category}     & -         & -    & $68.60$ & $56.51$ & $57.75$ & $55.54$ \\
 &      & -         & Random Crop & $68.81$ & $56.89$ & $58.05$ & $57.41$ \\
 &      & \ding{51} & -    & $68.06$ & $56.09$ & $58.65$ & $55.97$ \\
 &      & \ding{51} & Random Crop & $68.19$ & $56.78$ & $58.35$ & $57.12$ \\
 &      & \ding{51} & Multi-Margin & \textbf{68.94} & \textbf{57.30} & \textbf{59.81} & \textbf{57.63} \\
 \cline{2-8}
 & \multirow{5}{*}{Descriptions} & -         & -    & $72.67$ & $57.78$ & $61.61$ & $56.55$ \\
 &  & -         & Random Crop & $73.17$ & $58.87$ & $62.13$ & $57.99$ \\
 &  & \ding{51} & -    & $72.61$ & $58.70$ & $63.28$ & \textbf{59.35} \\
 &  & \ding{51} & Random Crop & $72.86$ & $58.99$ & $63.32$ & $58.78$ \\
 &  & \ding{51} & Multi-Margin & \textbf{73.49} & \textbf{59.34} & \textbf{64.05} & $59.06$ \\
 \hline
 \hline
 \parbox[t]{2mm}{\multirow{10}{*}{\rotatebox[origin=c]{90}{ViT-L/14}}}
 & \multirow{5}{*}{Category}     & -         & -    & $75.15$ & $63.08$ & $64.78$ & $62.16$ \\
&      & -         & Random Crop & $75.30$ & $63.32$ & $64.70$ & $62.59$ \\
 &      & \ding{51} & -    & $75.00$ & $62.96$ & $66.02$ & $62.16$ \\
 &      & \ding{51} & Random Crop & $75.04$ & $63.24$ & $66.54$ & $62.73$ \\
 &      & \ding{51} & Multi-Margin & \textbf{75.71} & \textbf{63.63} & \textbf{66.92} & \textbf{63.17} \\
 \cline{2-8}
 & \multirow{5}{*}{Descriptions} & -         & -    & $78.48$ & $64.65$ & $67.78$ & $63.17$ \\
 &  & -         & Random Crop & $78.65$ & $64.60$ & $67.65$ & \textbf{63.96} \\
 &  & \ding{51} & -    & $78.32$ & $64.67$ & $69.07$ & $63.31$ \\
 &  & \ding{51} & Random Crop & $78.28$ & \textbf{64.88} & $69.41$ & \textbf{63.96} \\
 &  & \ding{51} & Multi-Margin & \textbf{79.06} & $64.76$ & \textbf{69.88} & $62.95$ \\

\Xhline{4\arrayrulewidth}
\end{tabular}
\label{table:far_performance}
\end{table}

%% file: figures/tex/margin_vs_acc_far.tex
\begin{figure*}[tb]
    \centering
    \begin{subfigure}[b]{0.45\textwidth}
        \centering
        \includegraphics[width=\textwidth]{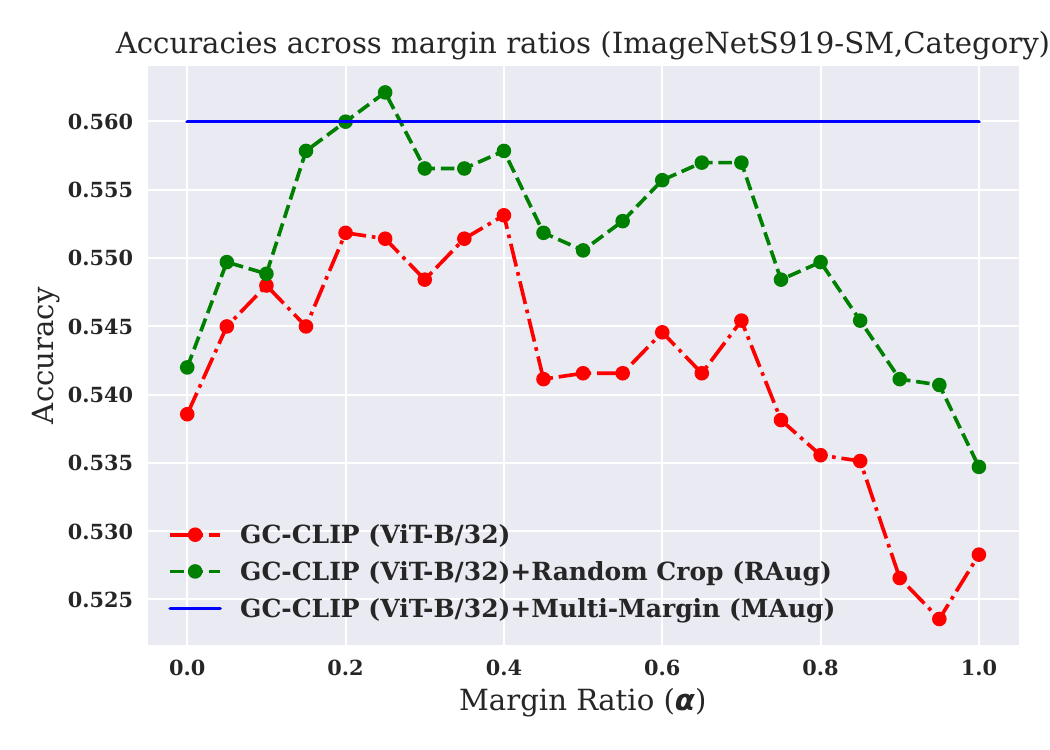}
        \caption{Prompt: Category}
        \label{fig:margin_vs_acc_far_imagenets919_single}
    \end{subfigure}
    \begin{subfigure}[b]{0.45\textwidth}
        \centering
        \includegraphics[width=\textwidth]{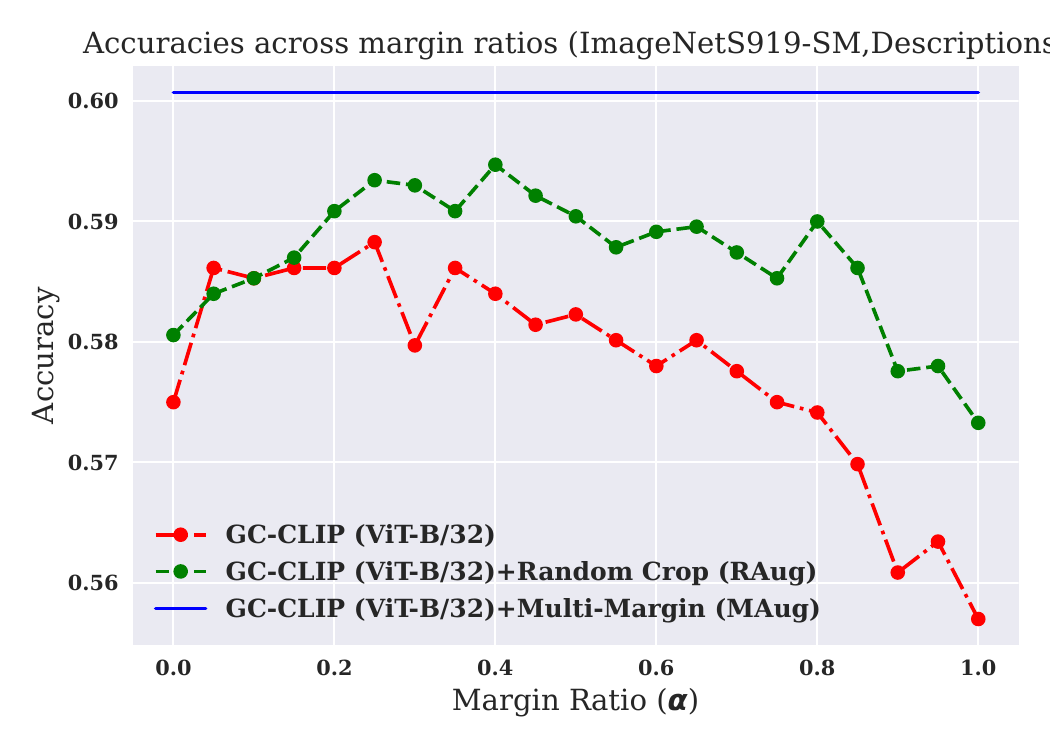}
        \caption{Prompt: Descriptions}
        \label{fig:margin_vs_acc_far_imagenets919_multi}
    \end{subfigure}

    \caption{Zero-shot accuracies on ImageNetS919-SM evaluated with different margin ratios.}
    \label{fig:margin_vs_acc_far}
\end{figure*}


%% file: figures/tex/object_size_vs_acc_vitb32.tex
\begin{figure*}[tb]
    \centering
    \begin{subfigure}[b]{0.45\textwidth}
        \centering
        \includegraphics[width=\textwidth]{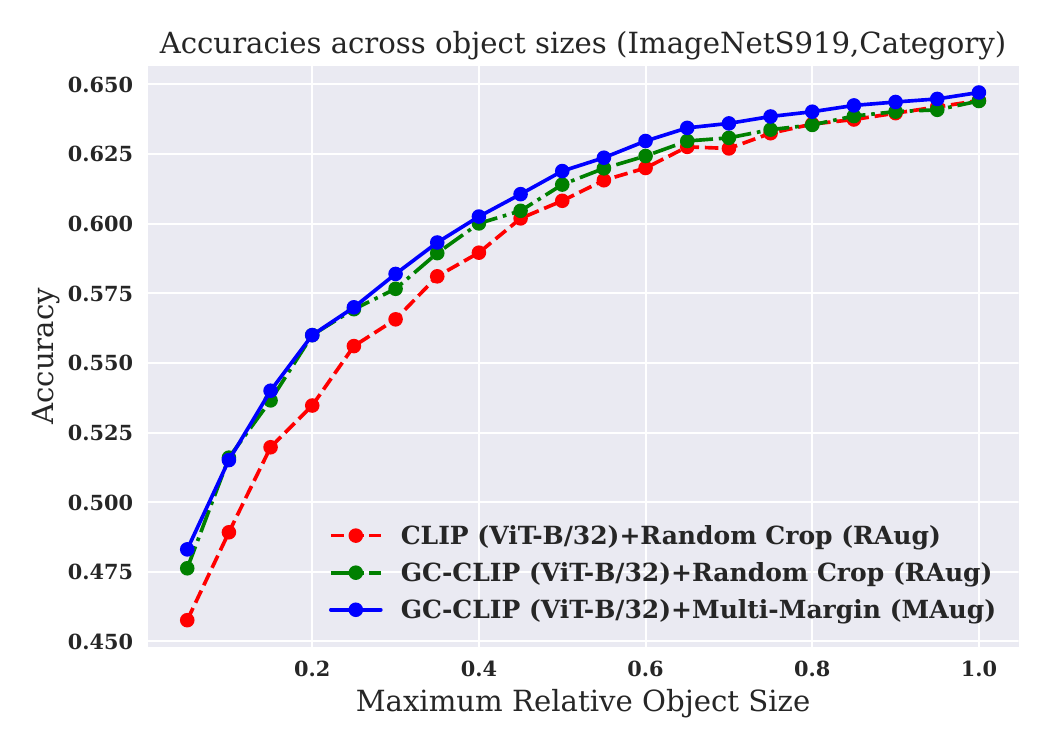}
        \caption{Prompt: Category}
        \label{fig:object_size_vs_acc_single_vitb32}
    \end{subfigure}
    \begin{subfigure}[b]{0.45\textwidth}
        \centering
        \includegraphics[width=\textwidth]{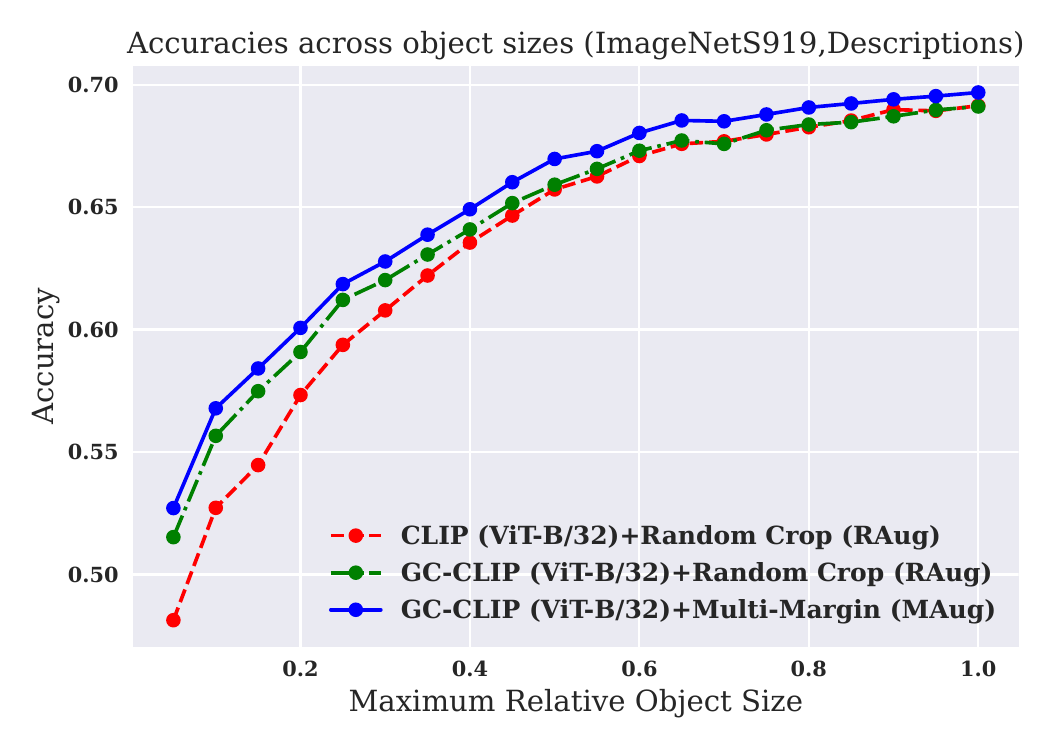}
        \caption{Prompt: Descriptions}
        \label{fig:object_size_vs_acc_multi_vitb32}
    \end{subfigure}

    \caption{Accuracies (ViT-B/32) on subsets of ImageNetS919 with various object size conditions.}
    \label{fig:object_size_vs_acc_vitb32}
\end{figure*}


%% file: figures/tex/qualitative_visualization.tex
\begin{figure}[ht!]
    \centering
    \begin{subfigure}[b]{0.38\textwidth}
        \centering
        \includegraphics[width=\textwidth]{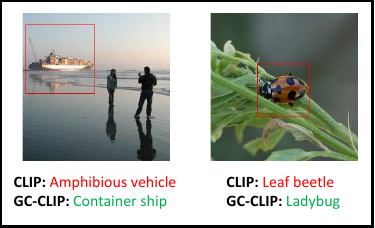}
        \caption{Improved cases}
        \label{fig:qualitative_visualization_good}
    \end{subfigure} \hspace{2cm}
    \begin{subfigure}[b]{0.38\textwidth}
        \centering
        \includegraphics[width=\textwidth]{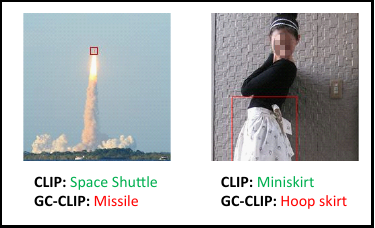}
        \caption{Failure cases}
        \label{fig:qualitative_visualization_bad}
    \end{subfigure}
    
    \caption{Predictions of CLIP (with RAug) and GC-CLIP (with MAug) with ViT-B/32 on ImageNetS919 samples. Red boxes represent primary boxes \(b^{0}\) estimated from our GC-CLIP.}
    \label{fig:qualitative_visualization}
\end{figure}

%% file: figures/tex/owl_incorrect_samples.tex
\begin{figure*}[tb]
    \centering
    \includegraphics[width=0.55\textwidth, height=0.20\linewidth]{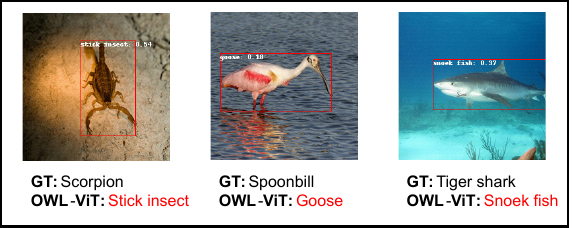}
    \caption{Examples of failure modes of the OWL-ViT based classifier.}
    \label{fig:owl_incorrect_samples}
\end{figure*}

%% file: sections/05_conclusion.tex
\section{Conclusion}

In this work, we identify a limitation of CLIP in zero-shot closed-set object classification task. As its image encoder is designed for encoding generic image representation, it is prone to encode non-discriminative context information into image features leading to performance degradation, particularly for small objects. We propose GC-CLIP, an approach to reduce effects from potentially non-discriminative information based on object bounding boxes estimated from a zero-shot object detection model. We empirically demonstrate that our approach outperforms baselines especially in cases of image samples with small objects. On the basis of ablation studies, we analyze conditions in which our approach performs well. We hope this work shed a new light on the behavior of large-scale open-vocabulary models for classification and guide future research to improve these models.

%% file: sections/07_supplement.tex
\appendix

\section{Appendix}

\subsection{Constructing dataset variations with small objects}
\label{sec:appendix_small_dataset_construction}

\input{figures/tex/dataset_osize_visualization}
\input{figures/tex/far_osize_vs_cnt}

In section \ref{sec:exp}, we use datasets based on ImageNetS and CUB as well as their small object variations (e.g., ImageNetS-SM and CUB-SM). In this section, we provide more details how those small variations are constructed.

For each image sample, its object size is computed based on object bounding box. In case of CUB, the bounding box is obtained directly from available annotations. However, for ImageNetS, only its pixel-wise segmentation is provided. In this case, object bounding box can be extracted from the segmentation in terms of minimum and maximum coordinates along \(X\) and \(Y\) axes of object-labelled pixels.

Given an image \(x_i\) of size \(w \times w\) with the object bounding box represented in terms of minimum/maximum \(XY\) coordinates as (\(p^{X}_{min}, p^{X}_{max}, p^{Y}_{min}, p^{Y}_{max}\)), relative object size of the image \(s_{x_i}\) is the ratio between the area of object bounding box and the total image area which can be computed as follows:
\begin{equation}
    s_{x_i} = \frac{(p^{X}_{max}-p^{X}_{min})(p^{Y}_{max}-p^{Y}_{min})}{w^2}.
\end{equation} The value of \(s_{x_i}\) will be within the range of \([0, 1]\). Example images with different values of \(s_{x_i}\) are shown in Figure \ref{fig:dataset_osize_visualization}.

We use \(s_{x_i}\) of individual image samples to control object size characteristic of a dataset. In section \ref{sec:exp}, the datasets with small objects (i.e., ImageNetS919-SM and CUB-SM), are obtained by thresholding \(s_{x_i}\) of image samples such that that their values are not larger than 0.2. In section \ref{sec:exp_osize_conditions}, multiple thresholds of \(s_{x_i}\) are employed on the ImageNetS919 dataset in order to study behavior of our models on different object size conditions. These thresholds are distributed uniformly from 0.05 to 1.0 with the step size of 0.05. The number of samples in each of these object size conditions is presented in Figure \ref{fig:far_osize_vs_cnt}.

\subsection{Pretrained supervised models with Guided Cropping}
\label{sec:appendix_supervised}

\input{tables/supervised_performance}

In the main paper, we mainly focus on applying our Guided Cropping to zero-shot models, i.e., CLIP. We argue that Guided Cropping can be helpful in this case as image encoders of these models are designed to be generic so that they potentially encode non-discriminative information of input images. Theoretically, our Guided Cropping can be applied to non-zero-shot models as well. In this section, we study behaviors of Guided Cropping when it is integrated with pretrained supervised models. In this regard, we utilize ImageNet pretrained models with ViT-B/32, ViT-B/16 and ViT-L/16 backbones from timm \cite{timm}, a deep learning library. These models are evaluated on ImageNetS919 and ImageNetS919-SM datsets with/without Guided Cropping. The results are shown in Table \ref{table:supervised_performance}.

According to the results, optimal performance generally achieves with models without Guided Cropping or with Guided Cropping using large margin ratio, i.e., 0.8, whose crops already cover large context regions. We can observe this behavior even in the case of small objects (ImageNetS919-SM). These results indicate that, for these supervised models, unrelated contexts generally do not degrade classification performance. In contrast, these contexts even improve their performance. This observation is actually not new and has been discussed in shortcut learning literature \cite{geirhos2020shortcut} that supervisedly trained networks can take unintended visual cues (e.g., background, texture) as shortcuts to gain classification performance on in-distribution samples.

\subsection{Logit refinement on top-k predictions}
\label{sec:appendix_top_k}

\input{tables/clip_accuracy_top_k}

As per our method mentioned in section \ref{sec:method_owl_clip}, after computing preliminary logits from conventional CLIP, only top-k predictions are considered and refined with Guided Cropping. We choose \(k=5\) in this work. In this section, we will provide reasons why we adopt this top-k refinement strategy. Two main reasons are given below.

\begin{itemize}
    \item Potential Accuracy: We found that there is already high chances that the correct classes are among predicted top-5 classes. To demonstrate this, we analyze top-1, top-5 and top-10 accuracies of conventional CLIP in Table \ref{table:clip_accuracy_top_k}. According to the results, large accuracy gaps can be noticed between top-1 and top-5 accuracies (24.53\% for ImageNetS919 and 31.79\% for CUB). In other words, by considering only 5 classes for refinement with Guided Cropping, upper bounds of final accuracies are already high. It must be noted that, while this upper bound accuracies can be raised further by considering top-10 classes, the gains compared to top-5 classes are relatively small. This may not worth introducing additional computation to the pipeline. Therefore, we decide to perform Guided Cropping based on predicted top-5 classes in this work.

    \item Common Bounding Boxes: We notice that visual appearances of top-5 classes are relatively similar in most cases. OWL-ViT is also likely to produce similar boxes for these classes. This makes the use of common bounding boxes (e.g., the primary box \(b^0_i\) or the \(\alpha\)-margin box \(b^{\alpha}_i\)) among these classes reasonable. To illustrate this, considering each sample in Figure \ref{fig:score_plots_1} and \ref{fig:score_plots_2}, its primary box generally contains visual features which are (partially) similar to each top class making the box become a decent box candidate for all top classes.
\end{itemize}

\subsection{Accuracies with different object size conditions}
\label{sec:appendix_osize_vs_acc}

\input{figures/tex/object_size_vs_acc_vitb16}
\input{figures/tex/object_size_vs_acc_vitl14}

In section \ref{sec:exp_osize_conditions}, we study GC-CLIP performance on various object size conditions and show that GC-CLIP variations outperform baselines especially when target object sizes are small. The plots in Figure \ref{fig:object_size_vs_acc_vitb32} are provided for models with ViT-B/32 backbone. In this section, additional evidences with other backbones are provided to support our claim. Figure \ref{fig:object_size_vs_acc_vitb16} and \ref{fig:object_size_vs_acc_vitl14} show similar plots for models with ViT-B/16 and ViT-L/14 backbones respectively. According to the figures, similar behavior can be observed. There are accuracy gaps between conventional CLIP and GC-CLIP and the gaps are larger on datasets with small objects. This demonstrates that our claim is consistent across different CLIP backbones.

\subsection{Inference with OWL-ViT}
\label{sec:appendix_object_detection_pass}

\input{tables/gc_clip_detection_pass}

OWL-ViT performs object detection taking images and text prompts as inputs and producing bounding boxes as well as their scores and class labels as outputs. In this work, for each image sample \(x_i\), we use OWL-ViT to extract bounding box candidates \(B_i\) based on a set of detection prompts of the top-k classes \(\left\{p^{det}_j | j \in J^{k}_i\right\}\). Theoretically, there are two possible options to obtain \(B_i\) from OWL-ViT.

\begin{itemize}
    \item Single Forward Pass (Single-Pass): with this option, an input image and all detection prompts are forwarded to OWL-ViT at once. With a single forward pass, OWL-ViT will produce a set of bounding boxes which will be used directly as \(B_i\).
    \item Multiple Forward Passes (Multi-Pass): with this option, OWL-ViT will perform forward pass with one detection prompt at a time. In other words, there will be \(k\) forward passes in total. Each forward pass will produce a set of bounding boxes \(b_{ij}\) based on a detection prompt \(p^{det}_j\). Bounding boxes estimated from all forward passes will be merged to get \(B_i\) according to equation \ref{eq:bounding_box_candidates}.
\end{itemize}

As mentioned in section \ref{sec:method_owl_clip}, we decide to adopt Multi-Pass in our Guided Cropping pipeline as Multi-Pass is more robust to misdetection (if one pass fails, other passes can act as backup passes). In this section, we demonstrate empirically that Multi-Pass can lead to better performance.

In this regard, we conduct an experiment to compare GC-CLIP accuracies when Single-Pass and Multi-Pass are employed. The results are shown in Table \ref{table:gc_clip_detection_pass}. According to the results, GC-CLIP with Multi-Pass is consistently better across datasets and model configurations. This confirms our design choice to use Multi-Pass in our Guided Cropping pipeline.

\subsection{Similarity between cropped images and their prompts}

\input{tables/similarity_score_change}

One motivation of our Guided Cropping is that, by minimizing unrelated information, CLIP image encoder can focus more on target objects leading to better image representations. In section \ref{sec:exp_overall_performance} better image representations can be indirectly inferred via the improvement of the classification performance. In this section, we would like to analyze image representations in another perspective.

We argue that, if image representations are better, the representations should be not only less similar to prompts of other classes but also more similar to prompts of their own classes. In this regard, we investigate similarities of image embeddings (of the correctly classified samples) to their own prompts. Here, similarity scores are obtained in terms of maximum predicted logit values. Similarity score results of CLIP and GC-CLIP are shown in Table \ref{table:similarity_score_change}. We can notice that similarity scores between images and their corresponding prompts in case of GC-CLIP are consistently higher. This indicates that image representations after Guided Cropping are more similar to their prompts according to our assumption.

\subsection{Visualizing example results}

\input{figures/tex/score_plots_1}
\input{figures/tex/score_plots_2}

In this section, we present top-5 logits estimated from CLIP and GC-CLIP on example samples from ImageNetS919 to demonstrate qualitatively that GC-CLIP can refine logits to make correct predictions. The results are illustrated in Figure \ref{fig:score_plots_1} and \ref{fig:score_plots_2}.


%% file: figures/tex/dataset_osize_visualization.tex
\begin{figure}[ht!]
    \centering
    \includegraphics[width=\textwidth]{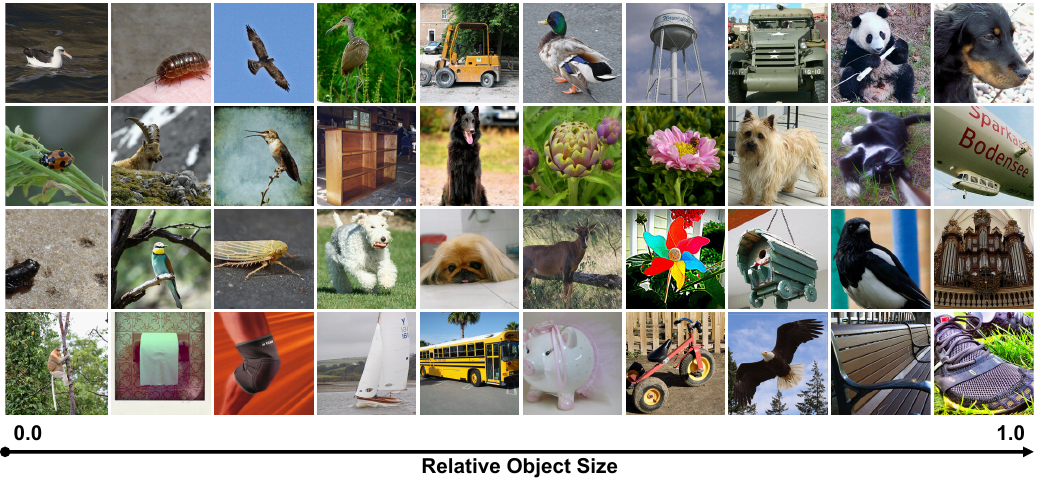}
    \caption{Example images from ImageNetS919 with different relative object sizes.}
    \label{fig:dataset_osize_visualization}

\end{figure}

%% file: figures/tex/far_osize_vs_cnt.tex
\begin{figure}[ht!]
    \centering
    \includegraphics[width=0.8\textwidth]{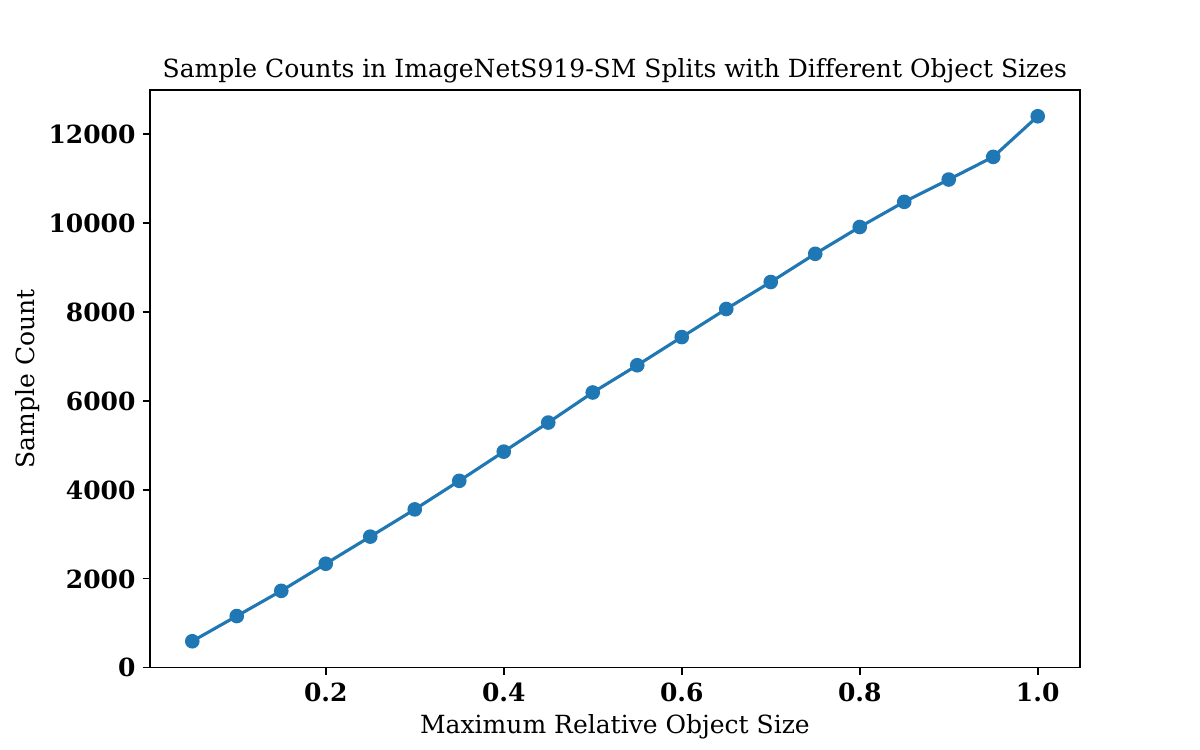}

    \caption{The number of samples in each object size condition of ImageNetS919.}
    \label{fig:far_osize_vs_cnt}
\end{figure}

%% file: tables/supervised_performance.tex
\begin{table}[t]

\caption{Classification accuracies of ImageNet pretrained models with/without Guided Cropping on ImageNet919.}

\centering
\begin{tabular}{c|c|c|c|c|c}
\Xhline{4\arrayrulewidth}
\multirow{2}{*}{Architecture} & \multirow{2}{*}{\makecell{Guided\\Cropping}} & \multirow{2}{*}{\makecell{Margin\\Ratio}} & \multirow{2}{*}{\makecell{Box\\Aug.}} & \multicolumn{2}{c}{Dataset} \\ 
 &&&& ImageNetS919 & ImageNetS919-SM \\ \hline
 ViT-B/32 & -         & -   & -    & $76.82$ & $61.53$ \\
 ViT-B/32 & -         & -   & Random Crop & $77.71$ & $62.21$ \\
 ViT-B/32 & \ding{51} & 0.2 & -    & $77.11$ & $64.05$ \\
 ViT-B/32 & \ding{51} & 0.2 & Random Crop & $77.99$ & \textbf{65.04} \\
 ViT-B/32 & \ding{51} & 0.8 & -    & $76.91$ & $62.81$ \\
 ViT-B/32 & \ding{51} & 0.8 & Random Crop & \textbf{78.14} & $63.84$ \\
 \hline
 ViT-B/16 & -         & -   & -    & $81.72$ & $68.89$ \\
 ViT-B/16 & -         & -   & Random Crop & \textbf{82.11} & \textbf{69.37} \\
 ViT-B/16 & \ding{51} & 0.2 & -    & $81.08$ & $68.42$ \\
 ViT-B/16 & \ding{51} & 0.2 & Random Crop & $81.16$ & $68.85$ \\
 ViT-B/16 & \ding{51} & 0.8 & -    & $81.63$ & $68.51$ \\
 ViT-B/16 & \ding{51} & 0.8 & Random Crop & $81.94$ & \textbf{69.37} \\
 \hline
 ViT-L/16 & -         & -   & -    & $86.09$ & $75.62$ \\
 ViT-L/16 & -         & -   & Random Crop & $86.35$ & \textbf{76.35} \\
 ViT-L/16 & \ding{51} & 0.2 & -    & $85.67$ & $75.92$ \\
 ViT-L/16 & \ding{51} & 0.2 & Random Crop & $85.69$ & $75.54$ \\
 ViT-L/16 & \ding{51} & 0.8 & -    & $86.21$ & $76.26$ \\
 ViT-L/16 & \ding{51} & 0.8 & Random Crop & \textbf{86.37} & \textbf{76.35} \\
\Xhline{4\arrayrulewidth}
\end{tabular}
\label{table:supervised_performance}
\end{table}



%% file: tables/clip_accuracy_top_k.tex
\begin{table}[t]

\caption{Top-k accuracies from conventional CLIP (ViT-B/32) with category prompts.}

\centering
\begin{tabular}{c|ccc}
\Xhline{4\arrayrulewidth}
\multirow{2}{*}{Dataset} & \multicolumn{3}{c}{Accuracy} \\ 
 & Top-1 & Top-5 & Top-10 \\ \hline
ImageNetS919 & 63.62 & 88.15 & 92.98 \\
CUB & 51.83 & 83.62& 90.63 \\
\Xhline{4\arrayrulewidth}
\end{tabular}
\label{table:clip_accuracy_top_k}
\end{table}

%% file: figures/tex/object_size_vs_acc_vitb16.tex
\begin{figure}[ht!]
    \centering
    \begin{subfigure}[b]{0.45\textwidth}
        \centering
        \includegraphics[width=\textwidth]{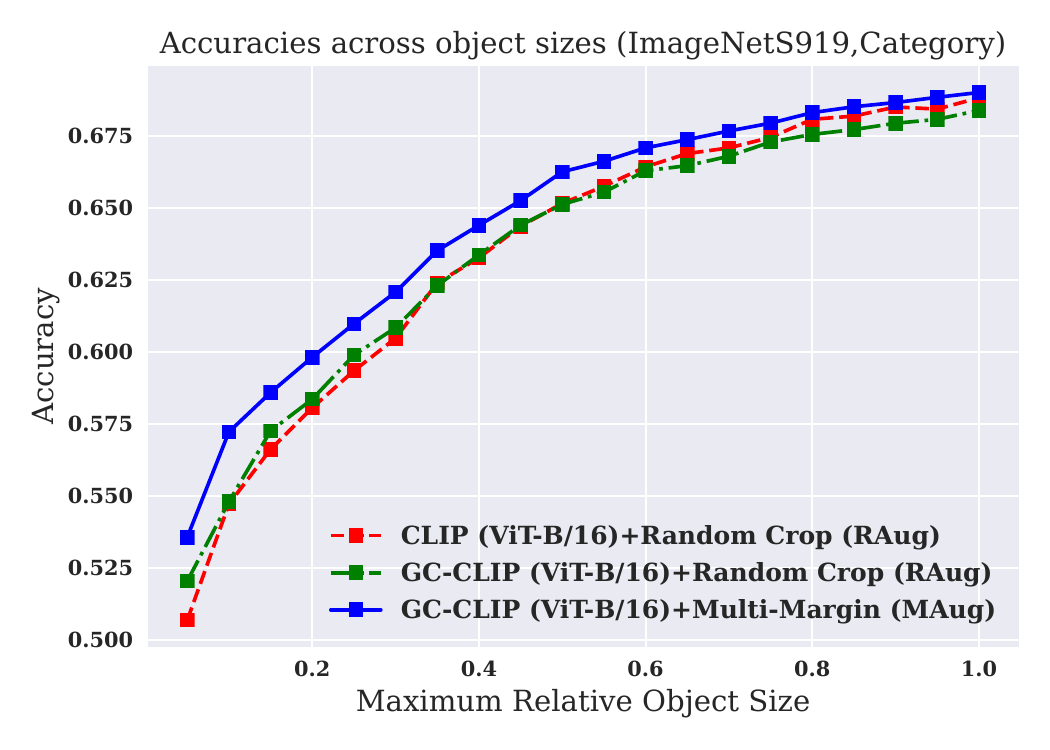}
        \caption{Prompt: Category}
        \label{fig:object_size_vs_acc_single_vitb16}
    \end{subfigure}
    \begin{subfigure}[b]{0.45\textwidth}
        \centering
        \includegraphics[width=\textwidth]{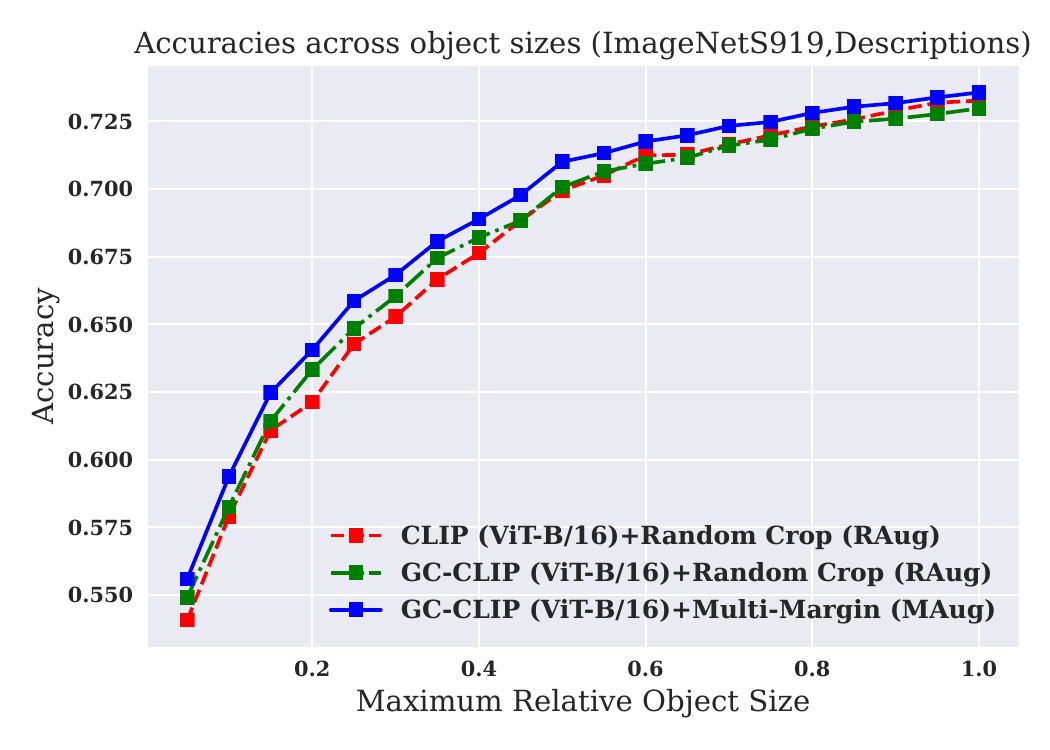}
        \caption{Prompt: Descriptions}
        \label{fig:object_size_vs_acc_multi_vitb16}
    \end{subfigure}

    \caption{Accuracies (ViT-B/16) on subsets of ImageNetS919 with various object size conditions.}
    \label{fig:object_size_vs_acc_vitb16}
\end{figure}


%% file: figures/tex/object_size_vs_acc_vitl14.tex
\begin{figure}[ht!]
    \centering
    \begin{subfigure}[b]{0.45\textwidth}
        \centering
        \includegraphics[width=\textwidth]{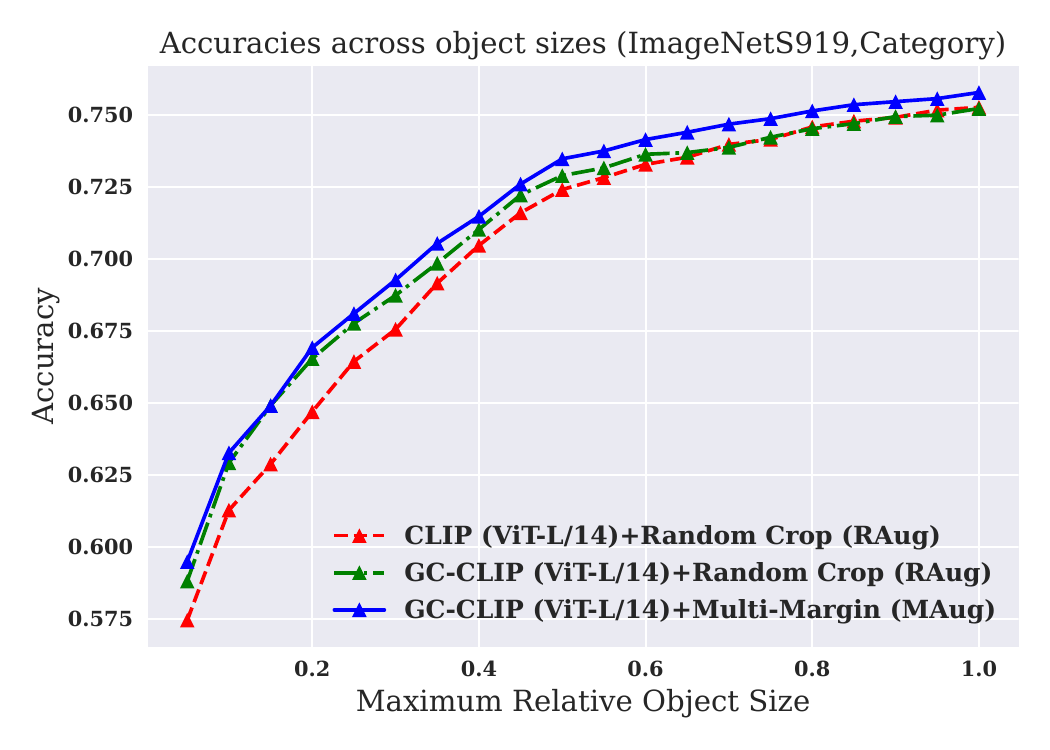}
        \caption{Prompt: Category}
        \label{fig:object_size_vs_acc_single_vitl14}
    \end{subfigure}
    \begin{subfigure}[b]{0.45\textwidth}
        \centering
        \includegraphics[width=\textwidth]{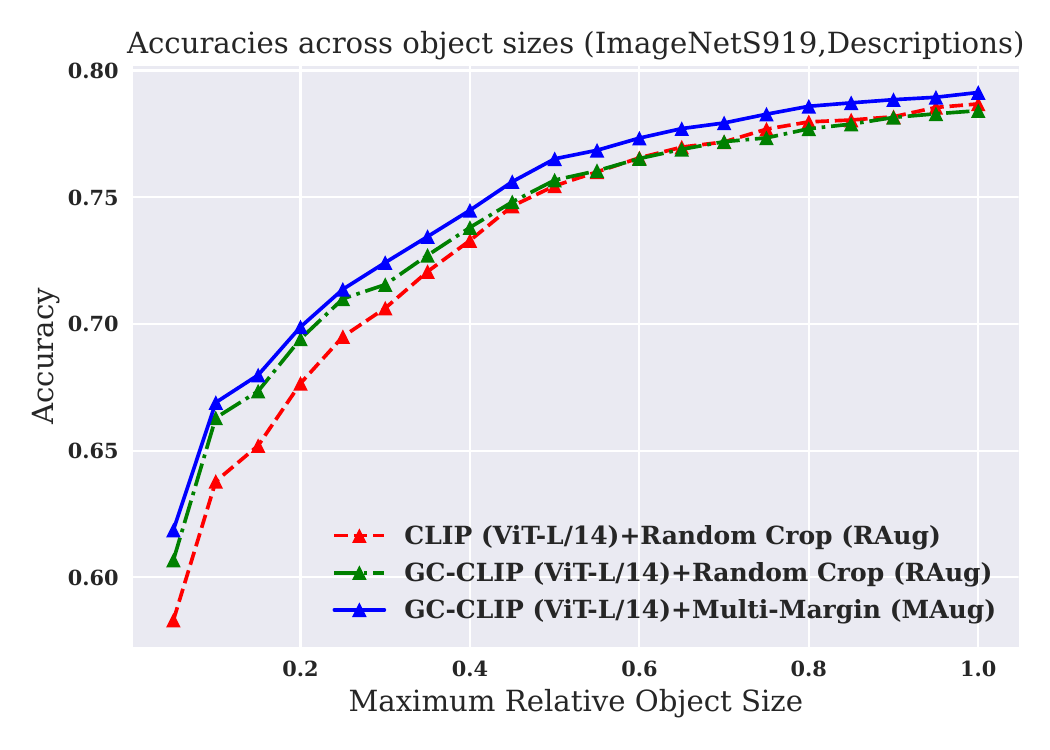}
        \caption{Prompt: Descriptions}
        \label{fig:object_size_vs_acc_multi_vitl14}
    \end{subfigure}

    \caption{Accuracies (ViT-L/14) on subsets of ImageNetS919 with various object size conditions.}
    \label{fig:object_size_vs_acc_vitl14}
\end{figure}


%% file: tables/gc_clip_detection_pass.tex
\begin{table}[t]

\caption{Accuracies from GC-CLIP (ViT-B/32) with different OWL-ViT inference strategies.}

\centering
\begin{tabular}{c|c|c||c|c}
\Xhline{4\arrayrulewidth}
\multirow{2}{*}{Dataset} & \multirow{2}{*}{Prompt Type} & \multirow{2}{*}{\makecell{Box\\Aug.}} & \multicolumn{2}{c}{OWL-ViT Inference} \\ 
 &&& Single-Pass & Multi-Pass \\ \hline
 ImageNetS919-SM & Category     & RAug & $54.71$ & \textbf{56.00} \\
 ImageNetS919-SM & Category     & MAug & $55.61$ & \textbf{56.00} \\
 \hline
 ImageNetS919-SM & Descriptions & RAug & $57.84$ & \textbf{59.08} \\
 ImageNetS919-SM & Descriptions & MAug & $59.47$ & \textbf{60.07} \\
 \hline
 CUB-SM          & Category     & RAug & $50.22$ & \textbf{52.81} \\
 CUB-SM          & Category     & MAug & \textbf{53.09} & \textbf{53.09} \\
 \hline
 CUB-SM          & Descriptions & RAug & $51.51$ & \textbf{53.09} \\
 CUB-SM          & Descriptions & MAug & \textbf{53.45} & $52.95$ \\

\Xhline{4\arrayrulewidth}
\end{tabular}
\label{table:gc_clip_detection_pass}
\end{table}

%% file: tables/similarity_score_change.tex
\begin{table}[t]

\caption{Average similarity scores between images and their corresponding prompts (i.e., maximum logit values) of correctly classified samples of CLIP (with RAug) and GC-CLIP (with MAug) using ViT-B/32 backbone.}

\centering
\begin{tabular}{c|c||c|c}
\Xhline{4\arrayrulewidth}
\multirow{2}{*}{Dataset} & \multirow{2}{*}{Prompt Type} & \multicolumn{2}{c}{Accuracy with } \\ 
 && CLIP & GC-CLIP \\ \hline
 ImageNetS919-SM & Category & 29.39 & \textbf{29.71} \\
 ImageNetS919-SM & Descriptions & 30.17 & \textbf{30.51} \\
 CUB-SM & Category & 33.71 & \textbf{33.89} \\
 CUB-SM & Descriptions & 34.30 & \textbf{34.55} \\

\Xhline{4\arrayrulewidth}
\end{tabular}
\label{table:similarity_score_change}
\end{table}

%% file: figures/tex/score_plots_1.tex
\begin{figure}[t]
    \centering
    
    \begin{subfigure}[b]{\textwidth}
        \centering
        \includegraphics[width=\textwidth]{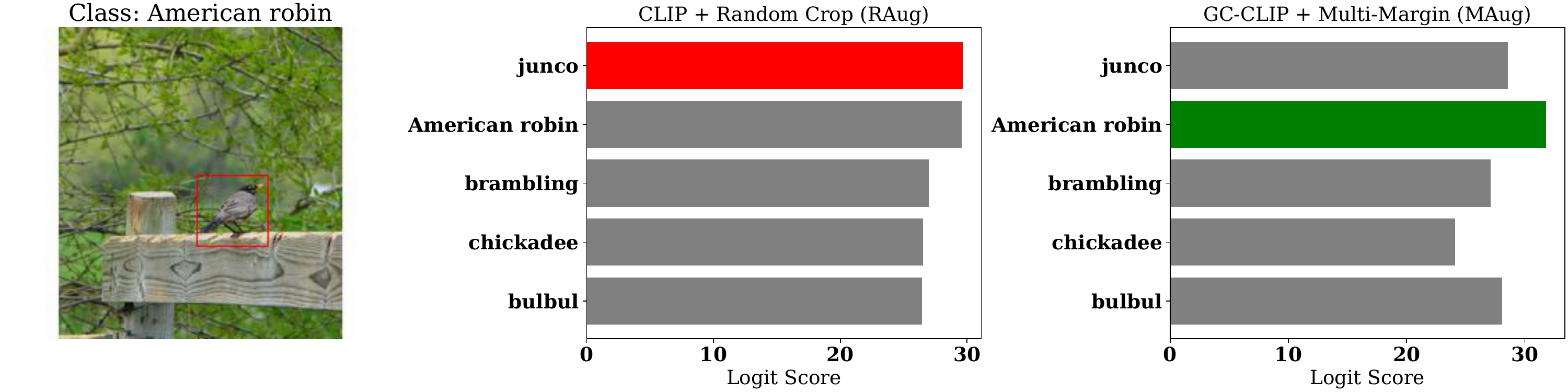}        
    \end{subfigure}
    \par\bigskip
    \begin{subfigure}[b]{\textwidth}
        \centering
        \includegraphics[width=\textwidth]{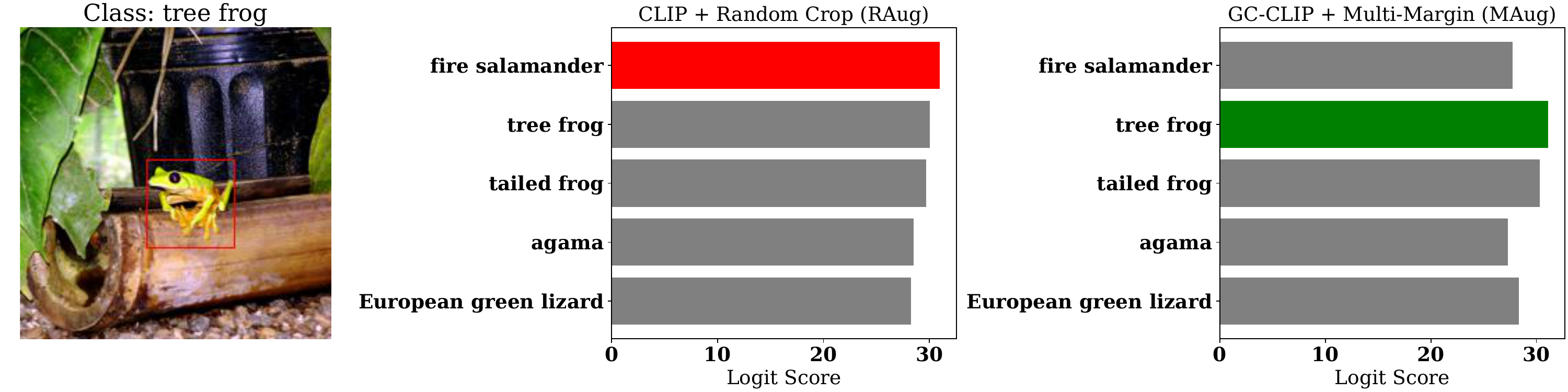}        
    \end{subfigure}
    \par\bigskip
    \begin{subfigure}[b]{\textwidth}
        \centering
        \includegraphics[width=\textwidth]{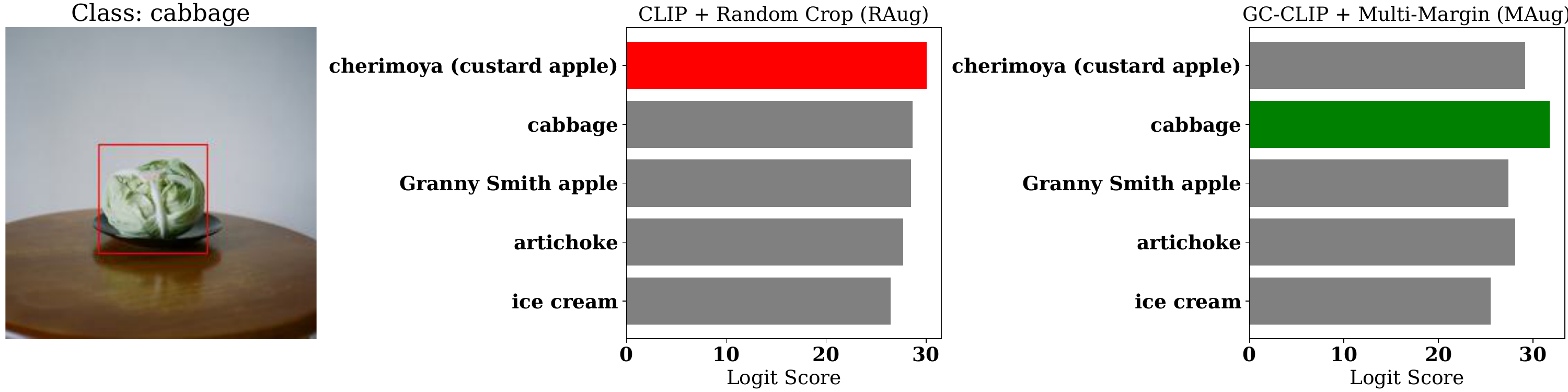}        
    \end{subfigure}    
    \par\bigskip
    \begin{subfigure}[b]{\textwidth}
        \centering
        \includegraphics[width=\textwidth]{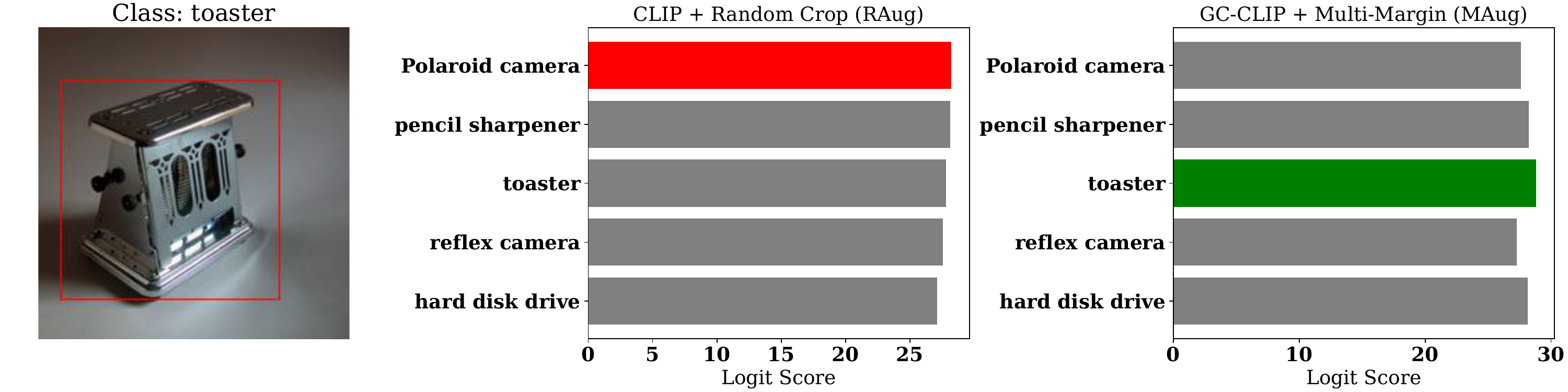}        
    \end{subfigure}    
    \par\bigskip
    \begin{subfigure}[b]{\textwidth}
        \centering
        \includegraphics[width=\textwidth]{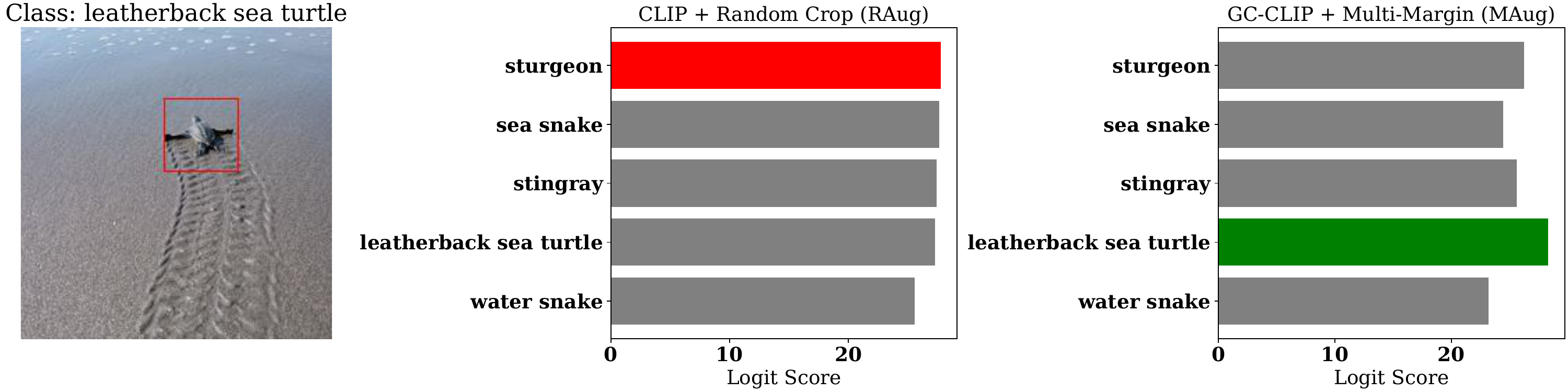}        
    \end{subfigure}        
    \caption{Top-5 logits on example samples improved by Guided Cropping (set 1). Model configurations are CLIP (with RAug) and GC-CLIP (with MAug) using ViT-B/32 backbone and prompt type of descriptions. Red boxes represent primary boxes used in our GC-CLIP pipeline.}
    \label{fig:score_plots_1}
\end{figure}

%% file: figures/tex/score_plots_2.tex
\begin{figure}[t]
    \centering
    
    \begin{subfigure}[b]{\textwidth}
        \centering
        \includegraphics[width=\textwidth]{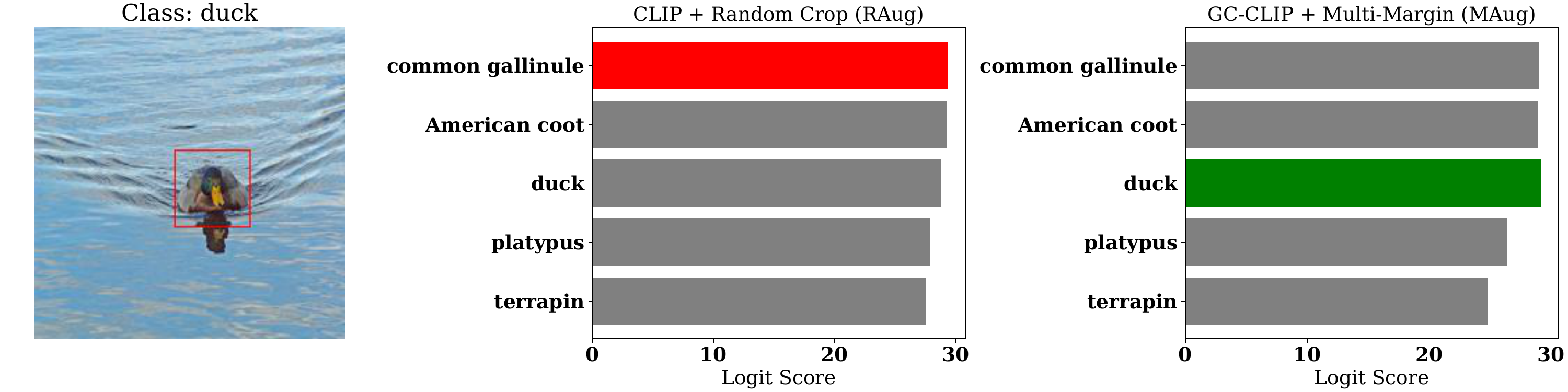}        
    \end{subfigure}
    \par\bigskip
    \begin{subfigure}[b]{\textwidth}
        \centering
        \includegraphics[width=\textwidth]{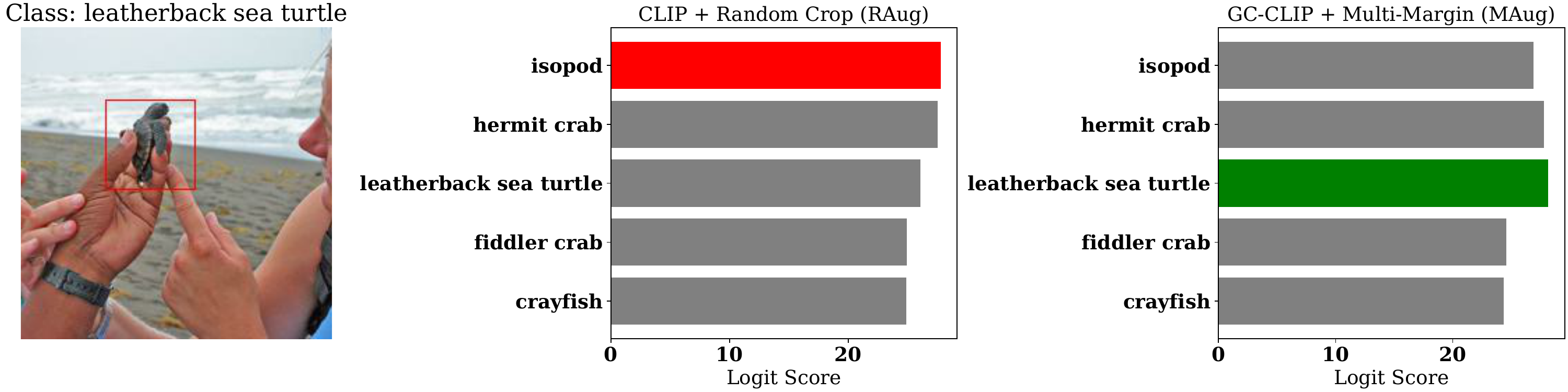}        
    \end{subfigure}
    \par\bigskip
    \begin{subfigure}[b]{\textwidth}
        \centering
        \includegraphics[width=\textwidth]{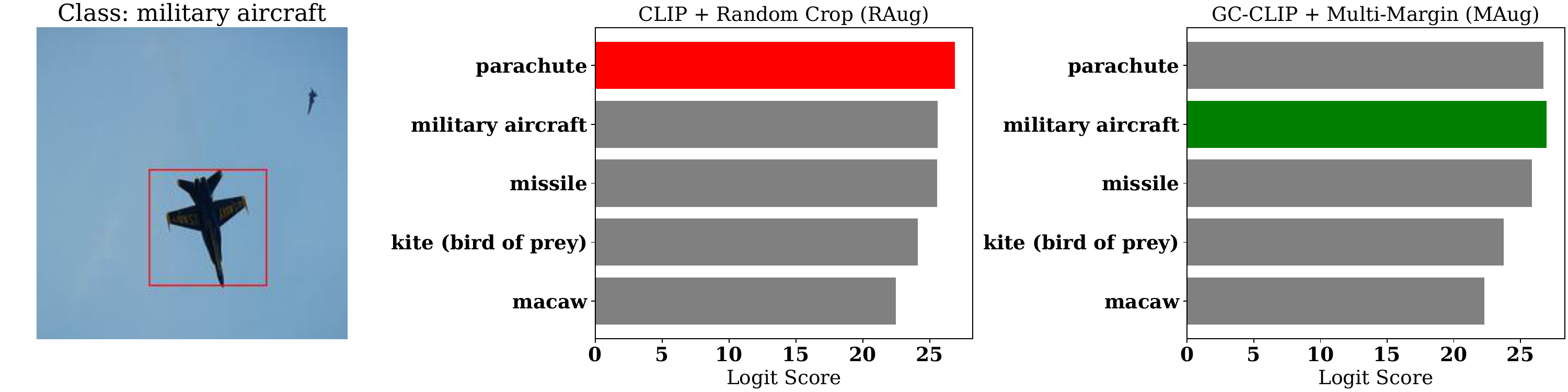}        
    \end{subfigure}    
    \par\bigskip
    \begin{subfigure}[b]{\textwidth}
        \centering
        \includegraphics[width=\textwidth]{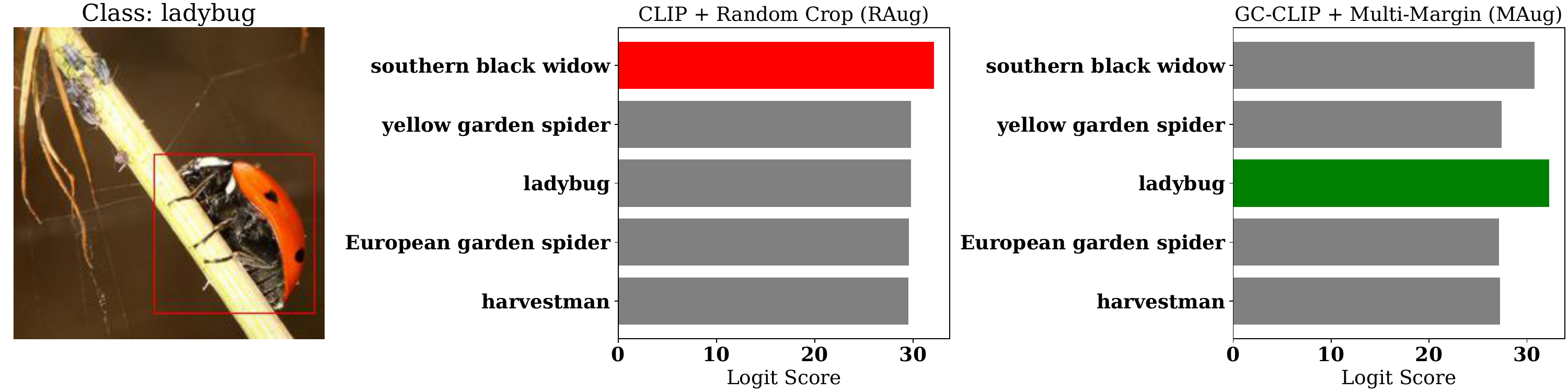}        
    \end{subfigure}    
    \par\bigskip
    \begin{subfigure}[b]{\textwidth}
        \centering
        \includegraphics[width=\textwidth]{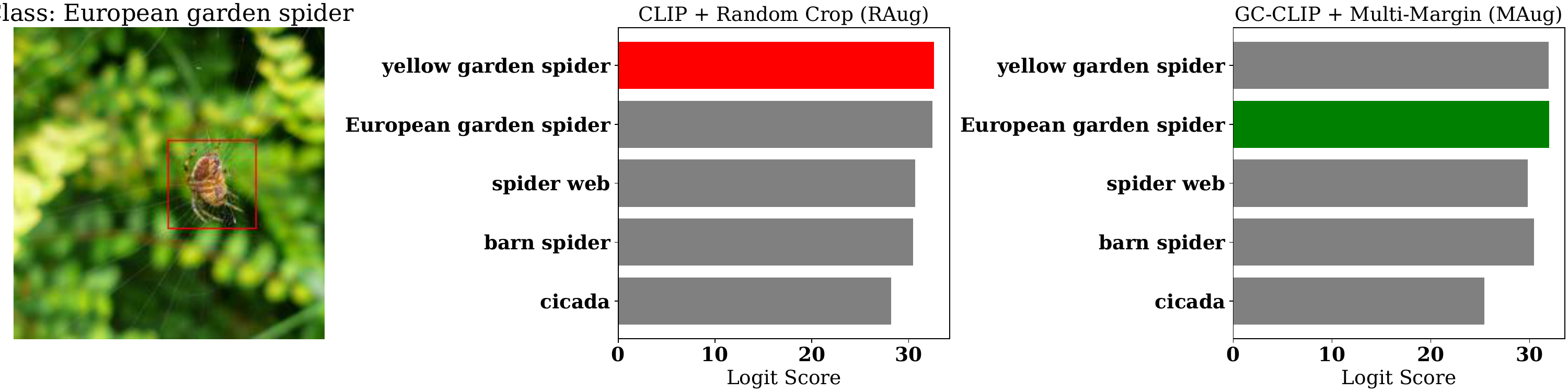}
    \end{subfigure}

    \caption{Top-5 logits on example samples improved by Guided Cropping (set 2). Model configurations are CLIP (with RAug) and GC-CLIP (with MAug) using ViT-B/32 backbone and prompt type of descriptions. Red boxes represent primary boxes used in our GC-CLIP pipeline.}
    \label{fig:score_plots_2}
\end{figure}

%% file: main.bbl
\begin{thebibliography}{10}

\bibitem{atzmon2020causal}
Yuval Atzmon, Felix Kreuk, Uri Shalit, and Gal Chechik.
\newblock A causal view of compositional zero-shot recognition.
\newblock In H.~Larochelle, M.~Ranzato, R.~Hadsell, M.~F. Balcan, and H.~Lin,
  editors, {\em Advances in Neural Information Processing Systems}, volume~33,
  pages 1462--1473. Curran Associates, Inc., 2020.

\bibitem{gao2022large}
Shanghua Gao, Zhong-Yu Li, Ming-Hsuan Yang, Ming-Ming Cheng, Junwei Han, and
  Philip Torr.
\newblock Large-scale unsupervised semantic segmentation.
\newblock {\em IEEE Transactions on Pattern Analysis and Machine Intelligence},
  2022.

\bibitem{geirhos2020shortcut}
Robert Geirhos, J{\"o}rn-Henrik Jacobsen, Claudio Michaelis, Richard Zemel,
  Wieland Brendel, Matthias Bethge, and Felix~A Wichmann.
\newblock Shortcut learning in deep neural networks.
\newblock {\em Nature Machine Intelligence}, 2(11):665--673, 2020.

\bibitem{gu2021open}
Xiuye Gu, Tsung-Yi Lin, Weicheng Kuo, and Yin Cui.
\newblock Open-vocabulary object detection via vision and language knowledge
  distillation.
\newblock {\em arXiv preprint arXiv:2104.13921}, 2021.

\bibitem{he2020deep}
Zhengyu He.
\newblock Deep learning in image classification: A survey report.
\newblock In {\em 2020 2nd International Conference on Information Technology
  and Computer Application (ITCA)}, pages 174--177. IEEE, 2020.

\bibitem{jia2021scaling}
Chao Jia, Yinfei Yang, Ye~Xia, Yi-Ting Chen, Zarana Parekh, Hieu Pham, Quoc Le,
  Yun-Hsuan Sung, Zhen Li, and Tom Duerig.
\newblock Scaling up visual and vision-language representation learning with
  noisy text supervision.
\newblock In {\em International Conference on Machine Learning}, pages
  4904--4916. PMLR, 2021.

\bibitem{kuo2022f}
Weicheng Kuo, Yin Cui, Xiuye Gu, AJ~Piergiovanni, and Anelia Angelova.
\newblock F-vlm: Open-vocabulary object detection upon frozen vision and
  language models.
\newblock {\em arXiv preprint arXiv:2209.15639}, 2022.

\bibitem{li2022grounded}
Liunian~Harold Li, Pengchuan Zhang, Haotian Zhang, Jianwei Yang, Chunyuan Li,
  Yiwu Zhong, Lijuan Wang, Lu~Yuan, Lei Zhang, Jenq-Neng Hwang, et~al.
\newblock Grounded language-image pre-training.
\newblock In {\em Proceedings of the IEEE/CVF Conference on Computer Vision and
  Pattern Recognition}, pages 10965--10975, 2022.

\bibitem{li2021learning}
Yong-Lu Li, Yue Xu, Xinyu Xu, Xiaohan Mao, and Cewu Lu.
\newblock Learning single/multi-attribute of object with symmetry and group.
\newblock {\em IEEE Transactions on Pattern Analysis and Machine Intelligence},
  2021.

\bibitem{mancini2021open}
Massimiliano Mancini, Muhammad~Ferjad Naeem, Yongqin Xian, and Zeynep Akata.
\newblock Open world compositional zero-shot learning.
\newblock In {\em Proceedings of the IEEE/CVF conference on computer vision and
  pattern recognition}, pages 5222--5230, 2021.

\bibitem{menon2022visual}
Sachit Menon and Carl Vondrick.
\newblock Visual classification via description from large language models.
\newblock {\em arXiv preprint arXiv:2210.07183}, 2022.

\bibitem{minderer2022simple}
Matthias Minderer, Alexey Gritsenko, Austin Stone, Maxim Neumann, Dirk
  Weissenborn, Alexey Dosovitskiy, Aravindh Mahendran, Anurag Arnab, Mostafa
  Dehghani, Zhuoran Shen, et~al.
\newblock Simple open-vocabulary object detection with vision transformers.
\newblock {\em arXiv preprint arXiv:2205.06230}, 2022.

\bibitem{naeem2021learning}
Muhammad~Ferjad Naeem, Yongqin Xian, Federico Tombari, and Zeynep Akata.
\newblock Learning graph embeddings for compositional zero-shot learning.
\newblock In {\em Proceedings of the IEEE/CVF Conference on Computer Vision and
  Pattern Recognition}, pages 953--962, 2021.

\bibitem{nagarajan2018attributes}
Tushar Nagarajan and Kristen Grauman.
\newblock Attributes as operators: factorizing unseen attribute-object
  compositions.
\newblock In {\em Proceedings of the European Conference on Computer Vision
  (ECCV)}, pages 169--185, 2018.

\bibitem{purushwalkam2019task}
Senthil Purushwalkam, Maximilian Nickel, Abhinav Gupta, and Marc'Aurelio
  Ranzato.
\newblock Task-driven modular networks for zero-shot compositional learning.
\newblock In {\em Proceedings of the IEEE/CVF International Conference on
  Computer Vision}, pages 3593--3602, 2019.

\bibitem{radford2021learning}
Alec Radford, Jong~Wook Kim, Chris Hallacy, Aditya Ramesh, Gabriel Goh,
  Sandhini Agarwal, Girish Sastry, Amanda Askell, Pamela Mishkin, Jack Clark,
  et~al.
\newblock Learning transferable visual models from natural language
  supervision.
\newblock In {\em International conference on machine learning}, pages
  8748--8763. PMLR, 2021.

\bibitem{russakovsky2015imagenet}
Olga Russakovsky, Jia Deng, Hao Su, Jonathan Krause, Sanjeev Satheesh, Sean Ma,
  Zhiheng Huang, Andrej Karpathy, Aditya Khosla, Michael Bernstein, et~al.
\newblock Imagenet large scale visual recognition challenge.
\newblock {\em International journal of computer vision}, 115:211--252, 2015.

\bibitem{welinder2010caltech}
Peter Welinder, Steve Branson, Takeshi Mita, Catherine Wah, Florian Schroff,
  Serge Belongie, and Pietro Perona.
\newblock Caltech-ucsd birds 200.
\newblock Technical Report CNS-TR-2010-001, California Institute of Technology,
  2010.

\bibitem{timm}
Ross Wightman.
\newblock Pytorch image models (timm).
\newblock \url{https://timm.fast.ai}.
\newblock Accessed: 2023-05-19.

\bibitem{yuan2021florence}
Lu~Yuan, Dongdong Chen, Yi-Ling Chen, Noel Codella, Xiyang Dai, Jianfeng Gao,
  Houdong Hu, Xuedong Huang, Boxin Li, Chunyuan Li, et~al.
\newblock Florence: A new foundation model for computer vision.
\newblock {\em arXiv preprint arXiv:2111.11432}, 2021.

\bibitem{zhang2022glipv2}
Haotian Zhang, Pengchuan Zhang, Xiaowei Hu, Yen-Chun Chen, Liunian Li, Xiyang
  Dai, Lijuan Wang, Lu~Yuan, Jenq-Neng Hwang, and Jianfeng Gao.
\newblock Glipv2: Unifying localization and vision-language understanding.
\newblock {\em Advances in Neural Information Processing Systems},
  35:36067--36080, 2022.

\bibitem{zhong2022regionclip}
Yiwu Zhong, Jianwei Yang, Pengchuan Zhang, Chunyuan Li, Noel Codella,
  Liunian~Harold Li, Luowei Zhou, Xiyang Dai, Lu~Yuan, Yin Li, et~al.
\newblock Regionclip: Region-based language-image pretraining.
\newblock In {\em Proceedings of the IEEE/CVF Conference on Computer Vision and
  Pattern Recognition}, pages 16793--16803, 2022.

\end{thebibliography}
